\documentclass{article}

\PassOptionsToPackage{numbers,compress}{natbib}
\usepackage[preprint]{neurips_2025}

\usepackage[T1]{fontenc}        
\usepackage[utf8]{inputenc}     

\usepackage{amsmath,amssymb,amsfonts}  
\usepackage{mathtools}                 
\usepackage{bm}                        
\usepackage{physics}                   
\usepackage{braket}                    

\usepackage[x11names,table]{xcolor}    
\usepackage{graphicx}                  
\usepackage{tikz}                      

\usepackage{booktabs}      
\usepackage{multirow}      
\usepackage{colortbl}      
\usepackage{dashrule}      
\usepackage{tablefootnote} 

\usepackage{algorithm}
\usepackage[noend]{algpseudocode}      
\usepackage[most]{tcolorbox}           

\usepackage{nicefrac}     
\usepackage{microtype}    
\usepackage{wrapfig}      
\usepackage{subcaption}   
\usepackage{placeins}     
\usepackage{enumerate}    
\usepackage{url}          

\usepackage{hyperref}
\hypersetup{
  colorlinks   = true,
  linkcolor    = RoyalBlue3,
  urlcolor     = RoyalBlue3,
  citecolor    = RoyalBlue3
}
\newcommand\firstpara[1]{\noindent\textbf{#1}\,}
\newcommand\para[1]{\noindent\textbf{#1}\,}

\definecolor{Green}{RGB}{0,155,85}
\definecolor{LightOrange}{rgb}{1,0.85,0.8}
\definecolor{LightPurple}{rgb}{1.0,0.80,0.95}
\definecolor{LightGreen}{rgb}{0.93,0.98,0.96}

\usepackage{wrapfig}
\usepackage{enumitem}

\title{ADLGen: Synthesizing Symbolic, Event-Triggered Sensor Sequences for Human Activity Modeling}

\author{
    \hspace{-1.0cm}\textbf{Weihang You$^{1}$\thanks{Equal contribution.}\,\,, %
    Hanqi Jiang$^{1}$\footnotemark[1]  , %
    Zishuai Liu$^{1}$, %
    Zihang Xie$^{2}$}\\
    \hspace{-1.0cm}\textbf{Tianming Liu$^{1}$, Jin Lu$^{1}$, Fei Dou$^{1}$\thanks{Corresponding author: fei.dou@uga.edu}}\\
    \hspace{-1.0cm}$^{1}$University of Georgia \quad
    $^{2}$University of Tsukuba \& Tokyo Institute of Technology \\[0.6em]
    \url{https://github.com/wyou-uga/ADLGen}\\
}

\begin{document}

\maketitle

\begin{abstract}
Real‑world collection of Activities of Daily Living (ADL) data is challenging due to privacy concerns, costly deployment and labeling, and the inherent sparsity and imbalance of human behavior. We present \textbf{ADLGen}, a generative framework specifically designed to synthesize realistic, event-triggered, and symbolic sensor sequences for ambient assistive environments. ADLGen integrates a decoder‑only Transformer with sign-based symbolic-temporal encoding,  and a context- and layout-aware sampling mechanism to guide generation toward semantically rich and physically plausible sensor event sequences. To enhance semantic fidelity and correct structural inconsistencies, we further incorporate a large language model (LLM) into an automatic generate–evaluate–refine loop, which verifies logical, behavioral, and temporal coherence and generates correction rules without manual intervention or environment-specific tuning. 
Through comprehensive experiments with novel evaluation metrics, ADLGen is shown to outperform baseline generators in statistical fidelity, semantic richness, and downstream activity recognition—offering a scalable and privacy-preserving solution for ADL data synthesis. 
\end{abstract}

\section{Introduction}
\label{sec:intro}

Activities of Daily Living (ADLs)—such as cooking, toileting, eating, and medication intake—are essential indicators of an individual's health status and autonomy \cite{arshad2022human, ye2024machine}. To unobtrusively monitor ADLs, Ambient Assisted Living (AAL) environments employ distributed non-wearable sensors throughout the home, including motion sensors, contact sensors (for doors/cabinets), pressure sensors (for bed/chair occupancy), and environmental sensors \cite{cicirelli2021ambient, Kang2023A}. These systems enable critical applications in elder care \cite{10658639, baig2019systematic}, early detection of cognitive impairment \cite{7470501, 10.3389/fbuil.2020.560497}, remote health monitoring \cite{gokalp2013monitoring, zhu2018sequence}, rehabilitation evaluation \cite{Andò2016A, Vermeulen2011Predicting}, and personalized behavioral interventions \cite{luperto2023integrating, stojchevska2024unlocking}. Compared to wearable or vision-based approaches, ambient sensing provides stronger privacy, user comfort, and long-term maintenance, aligning with the healthcare trend toward passive, preventive, and personalized care \cite{elhady2020sensor}.

These sensors capture human behavior as event-triggered sensor streams, shown in table \ref{tab:rawadldata}, where each event corresponds to a discrete sensor activation—such as a motion detector turning on, or a cabinet door being opened—along with a timestamp and sensor ID, and is recorded only when a status change occurs. For example, an event such as (2010-11-04, 09:34:16, D031, OPEN) may indicate that a laundry closet was opened. High-level activity annotations, such as COOKING or SLEEPING, are often manually labeled over time intervals and serve as semantic overlays to the raw event stream\cite{cook2012casas, alemdar2013aras,van2010activity_adldata}. These data exhibit several key characteristics: they are sparse and irregular (e.g. only recorded when a sensor is triggered), and symbolically discrete (e.g., ON/OFF, OPEN/CLOSE). They are also spatially grounded, as each sensor is tied to a specific room or object, and semantically layered, with low-level sensor activations composing higher-level human activity patterns. Public datasets such as the WSU CASAS corpus \cite{cook2012casas} exemplify this structure and are widely used benchmarks in the ADL modeling research \cite{DBLP:conf/cikm/JeonKYLK22,youhome2022,sreb-dataset,van2010activity}.

\begin{wrapfigure}[19]{R}{0.5\textwidth}
    \centering
    \includegraphics[width=0.5\textwidth]{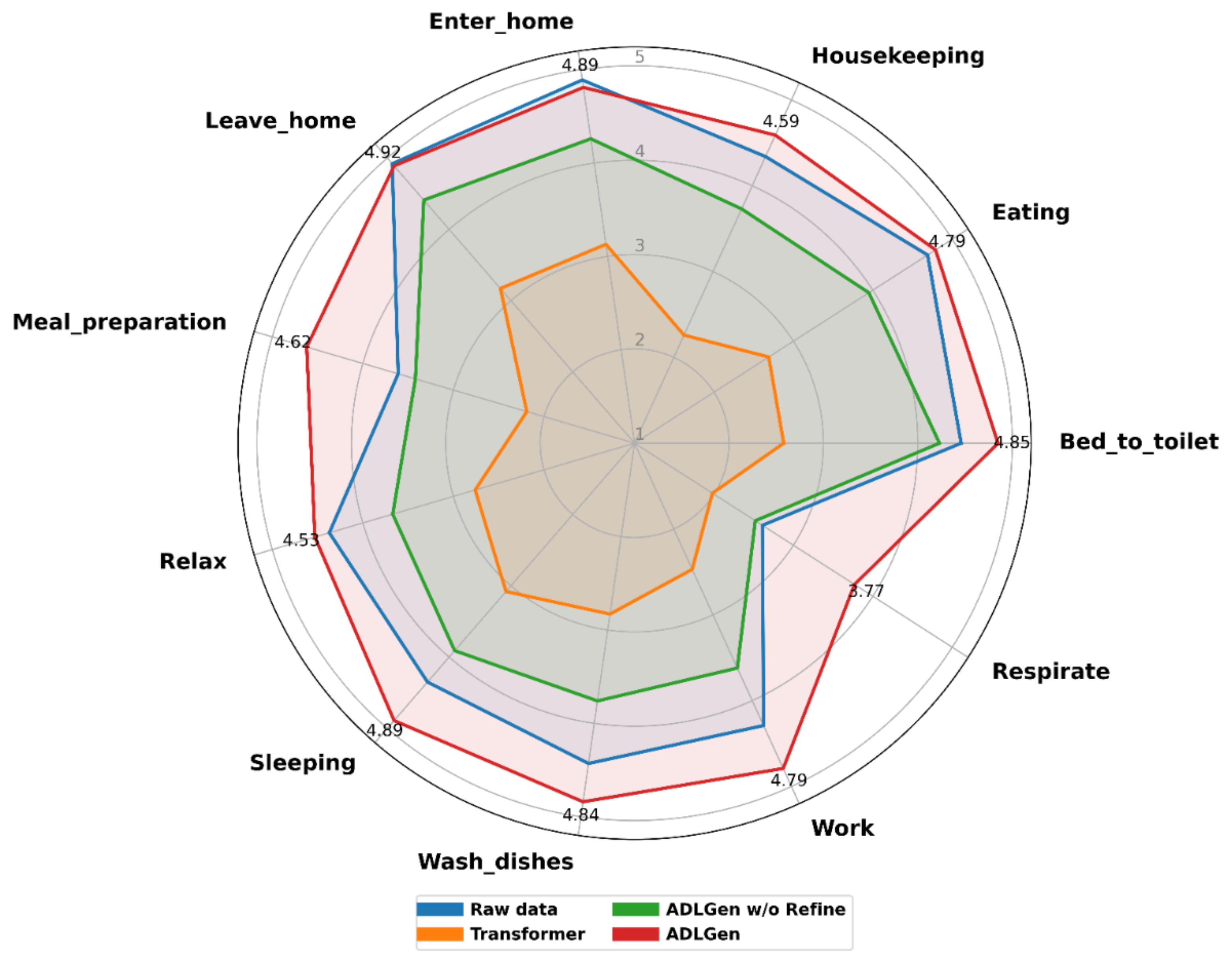}
    \caption{\small Semantic Quality evaluation (1-5 scale)\\
    across different daily activities.}
    \label{fig:intro}
\end{wrapfigure}

Despite the clinical and commercial importance of ADL modeling, large-scale data collection remains constraint by three intertwined barriers.  First, continuous monitoring in private spaces raises serious privacy concerns~\cite{mendes2025synthetic,lee2019living}.  Second, installing and maintaining dense sensor networks is time-consuming and costly and labor-intensive, especially for under-resourced facilities~\cite{chen2024enhancing_ieee}.  Third, the awareness of being monitored often alters natural behavior, compromising ecological validity~\cite{dsouza2025synthetic}. Consequently, existing datasets are limited in size, skewed in distribution, and lack sufficient coverage of rare but clinically important behaviors. Statistical fixes such as oversampling or class weighting~\cite{fernandez2018smote} merely rebalance labels without creating new, realistic behavior, and stronger privacy protection further degrade utility. Yet, while recognition systems have been widely studied, the potential of generative modeling to simulate realistic, high-fidelity ADL sequences remains unexplored---despite its potential to augment scarce data, preserve privacy, and support robust model training.

Recent advances in time series generation---using RNN/LSTM, GANs, diffusion models, VAEs, and transformers---have shown promise for continuous-valued sequences \cite{Zhao2023Multivariate, Ling2025RNDiff, Vuletić2024Fin-GAN}. However, these models fall short when applied to ADL sensor streams.
Most time series generators assume fixed-interval inputs, but aligning irregular, event-triggered ADL data to such formats often requires imputation that distorts temporal dynamics and disrupts activity rhythm
In addition, the symbolic discreteness of sensor activations is poorly aligned with models designed to generate smooth, continuous signals. Critically, these methods overlook the explicit spatial context linking sensor locations to human activities, which is vital for producing semantically valid and spatially realistic sequences.

To address these limitations, we introduce \textbf{ADLGen}, the \textit{first} generative framework specifically designed for synthesizing ADL sequences in ambient assistive environments. ADLGen tackles four core challenges of this domain: sparsity from event-triggered sampling, symbolic and discrete sensor state transitions, spatial grounding of sensor activations, and the need to preserve high-level activity semantics. It integrates a decoder-only Transformer with sign-based symbolic–temporal encoding, a context- and layout-aware sampling strategy, and an LLM-based semantic evaluation-and-refinement pipeline. Empirically, ADLGen produces sequences that closely match the statistical properties of real data, achieve superior semantic quality (Figure~\ref{fig:intro}), and significantly boost downstream recognition performance—offering a practical solution to both data scarcity and privacy constraints in ADL modeling.
In summary, the main contributions of our work include:
\begin{enumerate}[topsep=0pt,itemsep=0.3em,parsep=0pt,leftmargin=*]
    \item A novel decoder-only generative model with sign-based encoding and symbolic–temporal decoupling, which reduces sequence length, mitigates vocabulary inflation, and preserves semantic continuity in sparse, discrete, and heterogeneous event sequences. 
    \item 
          A two-stage context- and layout-aware sampling strategy that combines dynamic temperature modulation---adjusting randomness according to sequence length, diversity, and repetition---with spatial validation via sensor adjacency, ensuring that the generated sequences are both semantically diverse and physically plausible.
    \item 
        A hierarchical LLM-based evaluation-and-refinement pipeline that assesses sequences across logical consistency, behavioral coherence, and temporal regularity, and automatically generates correction rules without requiring site-specific expertise.
    \item A comprehensive performance evaluation protocol with novel metrics for statistical fidelity, diversity, physical plausibility, and semantic coherence—through which ADLGen is empirically shown to surpass all baselines in generation quality and downstream recognition tasks.
\end{enumerate}

\section{Related Work}
\label{sec:related_work}
The generation of diverse and realistic synthetic data for Activities of Daily Living (ADLs) is crucial to overcome the challenges of data scarcity, imbalance, and privacy inherent in data collection in the real world \cite{hwang2021eldersim, htun2024activity}. While ADLGen presents a novel end-to-end framework leveraging Transformers and LLMs for event-triggered sensor streams, its components and objectives relate to several active research areas.

\noindent \textbf{Data Encoding of Generative Models for Sequential and Event Data.} 
Recent advances in deep generative modeling span three primary approaches for temporal data: (1) Transformer-based models \cite{Transformer_Hawkes_ICML2020, SAHP_ICML2020} that handle irregular events through specialized attention mechanisms; (2) GAN-based frameworks \cite{yoon2019time, xu2020cotgan} that capture temporal dependencies via adversarial training; and (3) diffusion models \cite{DiffusionTS_NeurIPS2023, qian2024timeldm} that excel at continuous sequence generation through iterative denoising. Despite these advances, ADL data presents unique challenges due to its heterogeneous structure—combining discrete spatial identifiers (sensor IDs), binary states, and irregularly-sampled continuous timestamps \cite{thukral2025tdost, liciotti2019deepcasas,pan2020fine, marinov2022multimodal, petrich2022quantitative}. The encoding method adapted from time-series data that employ standard encoding strategies are suboptimal for ADL sequences, as they fail to preserve the spatial semantics of sensor activations and struggle with the mixed discrete-continuous nature of the data. Our work aims to address these limitations through proposed sign-based decoupling encoding approach, enabling more effective modeling of the unique characteristics of ADL sequences.

\noindent \textbf{Semantic and Field Knowledge-Driven Synthetic Data Generation.}
Ensuring semantic coherence and physical plausibility is crucial for synthetic Activities of Daily Living (ADL) data, which involves intricate logical sensor sequences, spatial constraints, and activity-specific patterns. Traditional approaches like SMOTE \cite{chawla2002smote} and even domain-constrained generative models \cite{khalafi2024constrained} struggle with ADL's complex spatial-temporal-semantic requirements \cite{Giannios2024A,Chatterjee2024Semantic}. While some prior work relies on manually crafted rules \cite{hemker2023cgxplain} or hard-coded spatial structures \cite{ahn2023star}, ADLGen introduces a novel LLM-driven generate-evaluate-refine pipeline. Unlike systems that use LLMs as direct sequence generators \cite{dai2025auggpt,latif2024systematic}, our approach employs LLMs as semantic validators and refinement agents \cite{Li2024Large,Liu2024Are}, leveraging their world knowledge to assess activity-specific coherence, generate corrective rules, and enhance low-frequency patterns \cite{Moncada-Ramirez2025Agentic,Zhan2025Leveraging}. This enables automated, knowledge-driven refinement without environment-specific expert intervention, addressing the nuanced logical ordering, spatial constraints, and activity-specific patterns essential for realistic ADL sequences.

\section{Method}
\label{sec:method}

\begin{figure}[t]
    \centering
    \includegraphics[width=\textwidth]{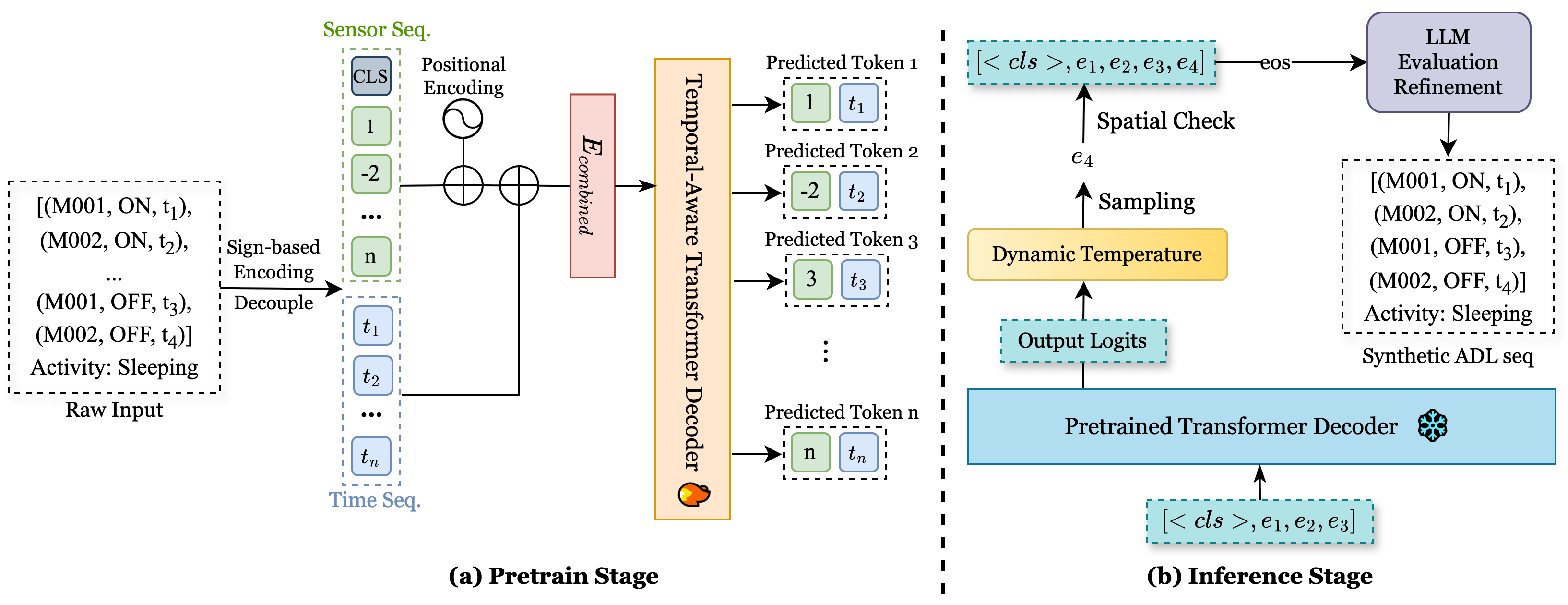}
    \caption{\textbf{Our Two-Stage Synthetic Activity Data Generation Framework.} (a) Pretrain Stage: Transformer learns ADL patterns from sign-based, decoupled encoded sensor data. (b) Inference Stage: Transformer generates sequences which are subsequently refined semantically by a large language model to enhance coherence and contextual relevance.}
    \label{fig:framework}
\end{figure}

ADLGen is an end-to-end framework for generating realistic event-triggered sensor streams of human activities. As shown in Figure \ref{fig:framework}, our approach integrates: (1) A decoder-only transformer with an LLM-driven evaluation and refinement pipeline. The transformer employs sign-based tokenization and temporal-symbolic decoupled representation to model ADL patterns, using context-aware sampling with spatial adjacency constraints during inference; (2) LLM-driven semantic evaluation and refinement framework: LLM-based semantic assessment that hierarchically evaluates ADL sequences across various designed perspectives. For sequences with violations, the LLM automatically generates corrective rules for refining them. ADLGen framework ensures that generated sequences match both statistical properties and semantic constraints of human activities, addressing the unique challenges of ADL data synthesis.

\subsection{Generative Framework for ADL Data Synthesis} \label{sec:gen_framework}

\firstpara{Problem Formulation.}
We formulate the ADL data synthesis as a conditional event sequence generation task, where the goal is to generate a sequence of realistic timestamped sensor activations $\boldsymbol{seq} = (e_1, e_2, \dots, e_M)$ with length $M$ ($M_{max}=100$ in our setting), conditioned on a high-level activity label $a \in \mathcal{A}$  (e.g., cooking). Each event $e_k$ is a symbolic tuple: $e_k = (\mathit{SensorID}_k, \mathit{State}_k, \mathit{Time}_k)$, where $\mathit{SensorID}_k \in \mathcal S$ identifies the triggered sensor, $\mathit{State}_k$ denotes its binary activation state---belonging to entiher $\{\text{ON},\text{OFF}\}$ or $\{\text{OPEN},\text{CLOSE}\}$ depending on the sensor type---and $\mathit{Time}_k$ records the exact timestamp of the event (see Table~\ref{tab:rawadldata} for an example). We model the conditional probability of a sequence $\boldsymbol{seq}$ given an activity $a$ as:
\begin{align}
p(\boldsymbol{seq}|a) = p(e_1|a) \textstyle\prod_{k=2}^{M} p(e_k | e_{<k}, a) \label{eq:generative_model}
\end{align}
where $e_{<k}\!\!=\!\!(e_1, e_2, \dots, e_{k-1})$ denotes the history of preceding generated events.
This formulation enables the model to synthesize semantically coherent, temporally consistent, and spatially grounded sensor event sequences that reflect real-world ADL patterns---offering a scalable and privacy-preserving alternative to data collection and a practical way to augment rare activity classes.


\begin{minipage}{0.36\textwidth}
    \centering
    \scriptsize
    \renewcommand{\arraystretch}{1}
     \setlength{\tabcolsep}{3pt}
    \captionsetup{type=table}
    \caption{\small Raw ADL Activity Data (Time: e.g., 2010-11-04,16:10:33.716795)} 
    \begin{tabular}{@{}l@{\hspace{3pt}}l@{\hspace{5pt}}l@{\hspace{5pt}}l@{\hspace{5pt}}l@{}}
    \toprule
    \textbf{Time} & \textbf{SensorID} & \textbf{State} & \textbf{Activity} & \textbf{Status} \\ %
    \midrule
    $t_1$ & M026 & ON & work & begin \\ 
    $t_2$ & M028 & OFF & & \\
    ... & ... & ...\\
    $t_{M-1}$ & M028 & ON & & \\
    $t_{M}$ & M026 & OFF & work & end \\ 
    \bottomrule
    \label{tab:rawadldata}
    \end{tabular}
\end{minipage}
\begin{minipage}{0.63\textwidth}
    \centering
    \scriptsize
    \renewcommand{\arraystretch}{1.3} 
     \setlength{\tabcolsep}{3pt}
    \captionsetup{type=table}
    \caption{\small Comparison of encoding strategies} 
    \begin{tabular}{@{}l@{\hspace{5pt}}l@{\hspace{5pt}}l@{\hspace{5pt}}l@{}}
    \toprule
    \textbf{Method} & \textbf{Example} & \textbf{Sensor Vocab} & \textbf{Tokens} \\
    \midrule
    Raw & [(M001, ON, $t_1$), (D001, CLOSE, $t_2$)] & N/A & $3M$ \\
    \midrule
    Flatten & [M001, ON, $t_1$, D001, CLOSE, $t_2$] & $|\mathcal{S}| + 4$ & $3M$ \\
    Composite & [M001\_ON, $t_1$, D001\_CLOSE, $t_2$] & $|\mathcal{S}| \times 2$ & $2M$ \\
    \textbf{Sign-based} & [+M001, -D001$] \& [t_1, t_2]$ & $|\mathcal{S}|$ & $M,M$ \\
    \bottomrule
    \label{tab:tokenization}
    \end{tabular}   
\end{minipage}\hfill

\paragraph{Generative Modeling of ADL Sensor Streams via Sign-Based Tokenization and Symbolic-Temporal Decoupling}
Tokenization strategies for ADL data can borrow from natural language processing by flattening structured event tuples into linear token streams. As shown in Table~\ref{tab:tokenization}, the \textit{flatten} approach interleaves sensor IDs, binary states, and timestamps as separate tokens, while the \textit{composite} approach concatenates sensor ID and state into compound tokens. 
While effective for natural language---where tokens carry compositional semantics---these approaches are ill-suited for structured, symbolic ADL  modeling. They introduce three key issues: (1) Sequence fragmentation, caused by alternating heterogeneous elements, disrupts structural locality for attention-based models; (2) Vocabulary inflation, as each sensor-state pair becomes a distinct token; (3) Semantic inconsistency, where the same sensor is mapped to unrelated embeddings under different states.
To address these limitations, we propose a \textit{sign-based tokenization} and \textit{symbolic-temporal decoupled encoding scheme} tailored to the structural and semantic characteristics of ADL event sequences.

Our sign-based representation encodes sensor ID and binary activation state into a single token $s_k$:
\begin{equation}
    s_k = Sign_k \cdot SensorID_k, \quad Sign_k \in \{-1,+1 \}
\end{equation}
where $+1$ denotes ON/OPEN states,  and $-1$ for OFF/CLOSED. This design offers multiple advantages: (1) vocabulary efficiency, as only sensor IDs form the vocabulary; (2) sequence compression without information loss; (3) compact encoding of spatial identity and activation state semantics; and (4) improved attention locality by eliminating interleaved ID-state-time token patterns.

We further decouple the symbolic and temporal components of each event to address the heterogeneous nature of ADL sequences. Instead of embedding (sign-based sensor, timestamp) as an interleaved stream, we split the input into two parallel sequences: a symbolic sequence $seq_{sens}\!\!=\!\![s_1, s_2, \dots, s_M]$ and a temporal sequence $seq_{time}\!\!=\!\![t_1, t_2, \dots, t_M]$. This decoupling helps the model capture meaningful contextual dependencies among symbolic activations—such as sensor firing patterns tied to specific activities—while separately encoding continuous, irregular timestamps using dedicated temporal embeddings. It improves representational clarity, reduces cross-modality interference, and enhances the model's ability to learn structured behavioral and temporal dynamics.


We integrates sensor semantics and temporal dynamics through a multi-component embedding:
\begin{align}
E(\boldsymbol{seq}_{sens},\boldsymbol{seq}_{temp}) &= E_{sens}(\boldsymbol{seq}_{sens}) + \omega_p \cdot PE(\boldsymbol{seq}_{sens}) + \omega_t \cdot E_{temp}(\boldsymbol{seq}_{temp}).
\end{align}

where $E_{sens}(\cdot)$ is the sign-based sensor token embeddings, $PE(\cdot)$ provides their positional context, and $E_{tmep}(\cdot)$ encodes temporal information using a time-aware embedding layer. The learnable coefficients $\omega_p$ and $\omega_t$ dynamically balance the sequential ordering and temporal influences. This formulation enables the model to adapt across ADL scenarios---from temporal cues in scheduled routines (e.g., medication) to symbolic patterns in flexible activities (e.g., relaxing).

To optimize this generative process, we minimize a composite loss function:
\begin{align}
\mathcal{L}_{total} = \mathcal{L}_{sensor_{id}}+\mathcal{L}_{sensor_{sign}} + \mathcal{L}_{temp} + \mathcal{L}_{special}
\end{align}
\noindent where $\mathcal{L}_{sensor_{id}}$ and $\mathcal{L}_{temp}$ are cross-entropy losses over predicted sensor IDs and timestamps, while $\mathcal{L}_{special}$ and $\mathcal{L}_{sensor_{sign}}$ are binary cross-entropy losses supervising special tokens and sensor activation state. 
This decomposed loss structure enables targeted optimization of the heterogeneous components in ADL sequences, allowing the model to simultaneously capture spatial, sequential, and temporal dependencies.

\para{Context- and Layout-Aware Iterative Sampling.}
\label{Context-Aware}
To generate realistic and semantically diverse ADL sequences, we adopt an iterative two-step sampling strategy during generation that jointly accounts for generation context and environmental layout. At each decoding step, the model first adjusts its sampling behavior based on the evolving sequence history, and then validates the physical feasibility of the proposed token using spatial constraints. This process comprises two components:

We dynamically adjust the sampling temperature based on the evolving characteristics of the sequence being generated. 
At each decoding step $t$, the final sampling distribution is computed as:
\begin{align}
logits_{final}(s_{t}) &= \frac{logits(s_t)}{\tau_{t}}, ~\tau_{t} = \frac{\tau_{base}\cdot R(\boldsymbol {seq}^t_{sens})}{\alpha(L_{t-1}) \cdot \beta(D_{s_t})}
\end{align}
$\tau_{t}$ is the proposed context-aware temperature, dynamically adjusted \textit{during generation} based on four contextual factors:(1) base temperature $\tau_{base}$ controlling randomness; (2) length-aware scaling factor $\alpha(L_t) = \tau_{min} + \frac{L_{t} - L_{min}}{L_{max} - L_{min}} \cdot (\tau_{max} - \tau_{min})$ encouraging exploration in early decoding stages and coherence later (temperature $\tau_{max/min}$ are hyperparameters, and $L_{min}=3, L_{max}=100$); (3) diversity factor $\beta(D_{s_t}) \in \{1,1.2\}$ increases temperature when sensor ID diversity $D_{s_t}$ is low; (4) repetition penalty $R(\boldsymbol {seq}^t_{sens})= {r_p \cdot (1.0 + 0.1 \cdot (n_{\boldsymbol {seq}^t_{sens}} - 1))}$ penalizes overused sensors while allowing natural repetitions, here, $n_{\boldsymbol {seq}^t_{sens}}$ is the frequency of the current sensor ID in generated sequence history, and $r_p$ is a tunable hyperparameter. This formulation allows temperature to evolve throughout the generation process, promoting diversity and flexibility in early stages and encouraging consistency and plausibility as the sequence unfolds.

To ensure that generated ADL sequences reflect realistic human movement, we constrain the generative process to produce only physically feasible transitions—i.e., transitions allowed by the spatial layout of the environment. This is implemented via floorplan-derived sensor adjacency matrix $A_{map} \in \{0, 1\}^{|\mathcal{S}| \times |\mathcal{S}|}$, where $A_{map}(i,j)=1$ indicates that sensors $i$ and $j$ are spatially connected (e.g., in the same room or linked through a walkable path). At each decoding step $t$, the model proposes a candidate sensor activation $s'_t$. We accept it only if it forms a valid transition with the previous token $s_{t-1}$ according to:
\begin{align}
s_{t} = 
\begin{cases} 
    s'_{t}, N=0 & \text{if~} A_{map}(|s'_{t}|, |s_{t-1}|) = 1 \\ 
    \text{RESAMPLE}, N+1 & \text{otherwise if~} N \leq 3
\end{cases}
\end{align}

If the proposed transition violates physical adjacency (i.e., it implies moving through walls or skipping disconnected regions), it is rejected and resampled. We allow up to \( N=3 \) times attempts to encourage diversity while maintaining physical realism. By enforcing these physically feasible transitions, our model avoids spatially implausible jumps, and ensures that generated sequences align with real-world environmental constraints.

\subsection{LLM-driven Semantic Evaluation and Refinement Framework} \label{sec:semantic_eval}

Traditional statistical metrics (e.g. MMD$^2$ \cite{gretton2006kernel_mmd}, diversity indices) fall short in evaluating the semantic and behavioral integrity of synthetic ADL sequences. We introduce a two-stage LLM-driven evaluation and refinement framework that leverages language models' semantic reasoning capabilities. Our method first converts symbolic sensor sequences into natural language descriptions using TDOST \cite{thukral2025tdost}, and then evaluates them using a multi-dimensional semantic analysis performed by a pretrained LLM. This allows LLMs to reason semantically over structured  event sequences without losing temporal or spatial context.
\subsubsection{LLM-based Semantic Evaluation}

\para{Hierarchical Semantic Quality Assessment.}
To evaluates the semantic plausibility of synthetic ADL sequences, we define a three-level hierarchy that captures distinct but interdependent dimensions of semantic quality: (1) fundamental logic ($\phi_F$) ensures physical feasibility (e.g., transitions between connected rooms) and logical consistency (e.g., no repeated 'ON' without 'OFF'); (2) behavioral coherence ($\phi_B$) assesses whether the sequence of sensor activations follows plausible action logic specific to the target activity, including room-function alignment (e.g., cooking occurs in the kitchen, not the bathroom), and action-level structure or order (e.g. door activation before room entry); and (3) temporal consistency ($\phi_T$) verifies that event timings conform to realistic temporal patterns associated with the given activity. 
Formally, we define:
\begin{align}
  \text{QualityScore}(\boldsymbol {seq}, a) = \phi_F(\boldsymbol {seq}) + \phi_B(\boldsymbol {seq}|a, \phi_F) + \phi_T(\boldsymbol {seq}|a, t, \phi_F)
\end{align}
This formulation enables principled, context-aware semantic evaluation while respecting the conditional dependencies across logic, physical, behavioral, and temporal reasoning dimensions.

\begin{wrapfigure}[13]{R}{0.68\textwidth}
    \centering
    \includegraphics[width=0.68\textwidth]{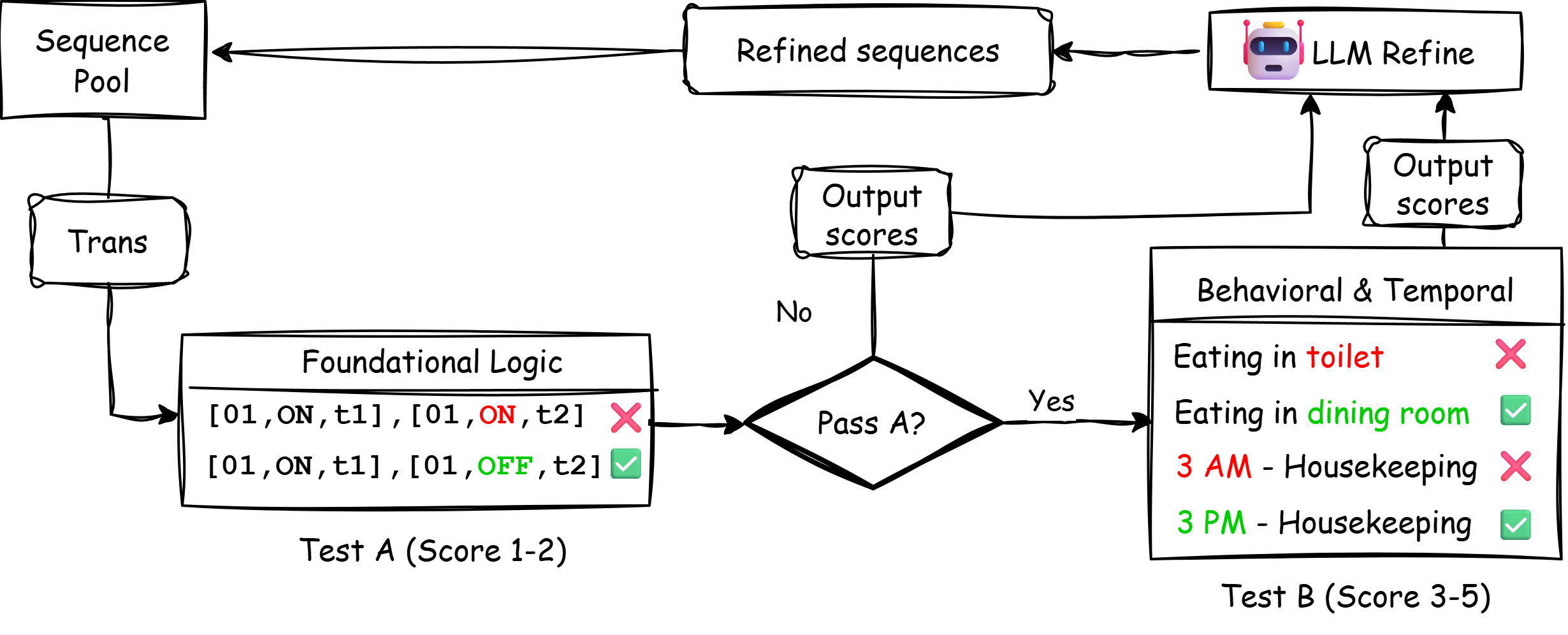}
    \caption{\small LLM evaluation framework with refinement pipeline.}
    \label{fig:llm_framework}
    \vspace{-15pt}
\end{wrapfigure}

\para{Two-Tiered Evaluation.}\\
The evaluation is organized in a two-tiered pipeline (Figure~\ref{fig:llm_framework}), inspired by Multi-Tiered Systems of Support (MTSS) \cite{stoiber2015mtss}. In Test A, the LLM validates fundamental logic by checking physical and logical consistency. Only sequences that pass Test A proceed to Test B, where the LLM jointly evaluates behavioral coherence and temporal consistency. This layered design allows efficient detection of diverse constraint violations, while distinguishing low- and high-level semantic errors.

\subsubsection{Refinement based on Semantic Evaluation}
For identified violations, the LLM automatically generates executable refinement rules. These rules are denoted as $\mathcal{R}=\{r_k=\text{LLM}(\boldsymbol{Seqs}, \Phi)\,|\,k=1,\dots,K\}$, where $\Phi$ provides the LLM context during semantic assessment. Each rule in $\mathcal{R}$ is designed to address a specific violation type of error within $\boldsymbol{Seqs}=\{\boldsymbol{seq_1}, \boldsymbol{seq_2},....\}$, resulting in a corrected sequence set $\boldsymbol{Seqs'}$.
The rule types include:
insertion rules adding missing logical preconditions (e.g., incorporating door sensor activation), deletion rules eliminating physically impossible events, and reordering rules rectifying temporal inconsistencies. Leveraging the LLM’s instruction-following capability, this refinement process is fully automated and generalizes across different environments without requiring task-specific rules or domain expertise. 

Please refer to Appendix for detailed description for LLM evaluation and refinement details.





\section{Empirical Results}
\label{sec:results}
\firstpara{Datasets and Experimental Setup.}
We evaluated ADLGen on the CASAS Aruba dataset \cite{cook2012casas}, which contains seven months of binary sensor data (motion, door) from a single-occupant apartment annotated with 12 ADLs. Some classes suffer from data insufficiency (e.g., only 33 'Housekeeping' and 6 'Respirate' samples), making this an ideal benchmark. Preprocessing included our sign-based sensor representation, temporal feature encoding (e.g. hour-of-day), and exclusion of "Other Activity" labels and temperature sensors to focus on explainable ADLs and events. All experiments used a consistent 5-fold cross-validation (90\% training, 10\% testing).
\subsection{Intrinsic Evaluation of Generated Sequences}
\label{sec:intrinsic}

\para{Setup.}
We evaluate ADLGen against several baselines: statistical resampling, LSTM, VQVAE, standard Transformer (all with our sign-based tokenization), and adapted cWGAN \cite{use_cWGAN} and Seq2Res \cite{chen2024generative}. Our evaluation employs four complementary aspects with the proposed novel metrics: (1) \textbf{Statistical fidelity} via Maximum Mean Discrepancy (MMD$^2$\,$\downarrow$) \cite{gretton2006kernel_mmd}; (2) \textbf{Diversity} through Intra-Set Similarity\,$\downarrow$ and entropy-based Diversity Score\,$\uparrow$; (3) \textbf{Physical plausibility} via Validity Rate\,$\uparrow$ using the sensor adjacency matrix from Section~\ref{Context-Aware}; and (4) \textbf{Semantic coherence} through LLM-assessed Semantic Quality (1-5)\,$\uparrow$ described in Section~\ref{sec:semantic_eval}. Detailed metric implementations are in the Appendix, with results summarized in Table~\ref{tab:gen_quality}.

Detailed implementations of these metrics are provided in the Appendix. Table~\ref{tab:gen_quality} summarizes the comparative performance across all evaluation dimensions.

\begin{table}[ht]
\vspace{-4pt}
\centering
\caption{\small Comparative Intrinsic Evaluation of Generative Models on ADL Sequence Quality. Our methods are highlighted, best results are in \textbf{bold}. 
}
\label{tab:gen_quality}
\small
\setlength{\tabcolsep}{3pt}
\renewcommand{\arraystretch}{0.8}
\begin{tabular}{l|cccc|c}
\toprule
\textbf{Generation Method} & \textbf{Realism} & \textbf{Intra-Set} & \textbf{Diversity} & \textbf{Validity} & \textbf{Semantic} \\
 & \textbf{(MMD$^2$↓)} & \textbf{Similarity$\downarrow$} & \textbf{Score$\uparrow$} & \textbf{Rate$\uparrow$} & \textbf{Quality$\uparrow$} \\
\midrule
Raw Data & N/A & 0.48 & 0.80 & 1.00 & 4.23 \\
\midrule
Statistical Resampling & 0.0286$\pm$0.0022 & 0.91$\pm$0.12 & 0.24$\pm$0.06 & 0.47$\pm$0.08 & 1.95$\pm$0.23 \\
LSTM & 0.0201$\pm$0.0031 & 0.68$\pm$0.15 & 0.55$\pm$0.06 & 0.54$\pm$0.07 & 2.16$\pm$0.26 \\
VQVAE & 0.0126$\pm$0.0019 & 0.62$\pm$0.16 & 0.63$\pm$0.05 & 0.57$\pm$0.05 & 2.41$\pm$0.24 \\
Transformer & 0.0118$\pm$0.0016 & 0.59$\pm$0.15 & 0.66$\pm$0.03 & 0.63$\pm$0.03 & 2.65$\pm$0.25 \\
cWGAN\cite{use_cWGAN} & 0.0095$\pm$0.0018 & 0.55$\pm$0.13 & 0.70$\pm$0.05 & 0.75$\pm$0.04 & 2.85$\pm$0.21 \\
Seq2Res\cite{chen2024generative} & 0.0088$\pm$0.0017 & 0.53$\pm$0.13 & 0.72$\pm$0.04 & 0.80$\pm$0.05 & 2.95$\pm$0.23 \\
\midrule
\rowcolor{LightGreen}ADLGen (w/o Refine) & 0.0047$\pm$0.0015 & 0.49$\pm$0.14 & 0.77$\pm$0.05 & 0.94$\pm$0.03 & 3.79$\pm$0.22 \\
\rowcolor{LightGreen}ADLGen (Full) & \textbf{0.0019$\pm$0.0009} & \textbf{0.41$\pm$0.11} & \textbf{0.86$\pm$0.03} & \textbf{1.00} & \textbf{4.67$\pm$0.22} \\
\bottomrule
\end{tabular}
\label{tab:gen_quality}
\end{table}

\para{Overall Performance.}
As shown in Table \ref{tab:gen_quality}, our full ADLGen framework establishes superior performance across all evaluated dimensions. It achieves the highest realism, best diversity, surpassing even raw data diversity—perfect physical plausibility, and the top Semantic Quality score. These results confirm ADLGen's ability to generate synthetic ADL sequences that are authentic, diverse, physically sound, and semantically rich.

\para{Impact of LLM Refinement.}
The LLM-based refinement stage is critical to ADLGen's success. While ADLGen (w/o Refine) already outperforms other generative models, the full ADLGen demonstrates substantial gains. Most notably, Semantic Quality increases from 3.79 to 4.67, exceeding the raw data's score. This suggests that the LLM refinement not only corrects flaws but actively enhances sequences to align with more complete and idealized representations of activities, improving upon inherent variabilities in real-world captures.

\para{Comparison with Baselines.}
Existing generative methods, from statistical resampling to more advanced models like cWGAN~\cite{use_cWGAN} and Seq2Res~\cite{chen2024generative}, exhibit clear limitations. Table \ref{tab:gen_quality} indicates that these baselines struggle to match ADLGen's low MMD$^2$ scores and high diversity. Furthermore, their physical validity often remains suboptimal. A crucial distinction lies in semantic quality: no baseline model surpasses a score of 3.0, highlighting their inadequacy in generating behaviorally coherent and contextually meaningful ADL sequences compared to ADLGen's specialized architecture and refinement pipeline.
\subsection{Extrinsic Evaluation: Behavioral Authenticity}
Following the strong intrinsic evaluation results, we further investigate the behavioral authenticity of ADLGen sequences by visualizing their distributional similarity to real data in classifier-learned feature spaces using t-SNE. Embeddings are extracted from an intermediate layer of a Bi-LSTM~\cite{zhou2016bilstm} activity recognition classifier.

\begin{wrapfigure}[21]{R}{0.55\textwidth}
    \centering
    \includegraphics[width=0.53\textwidth]{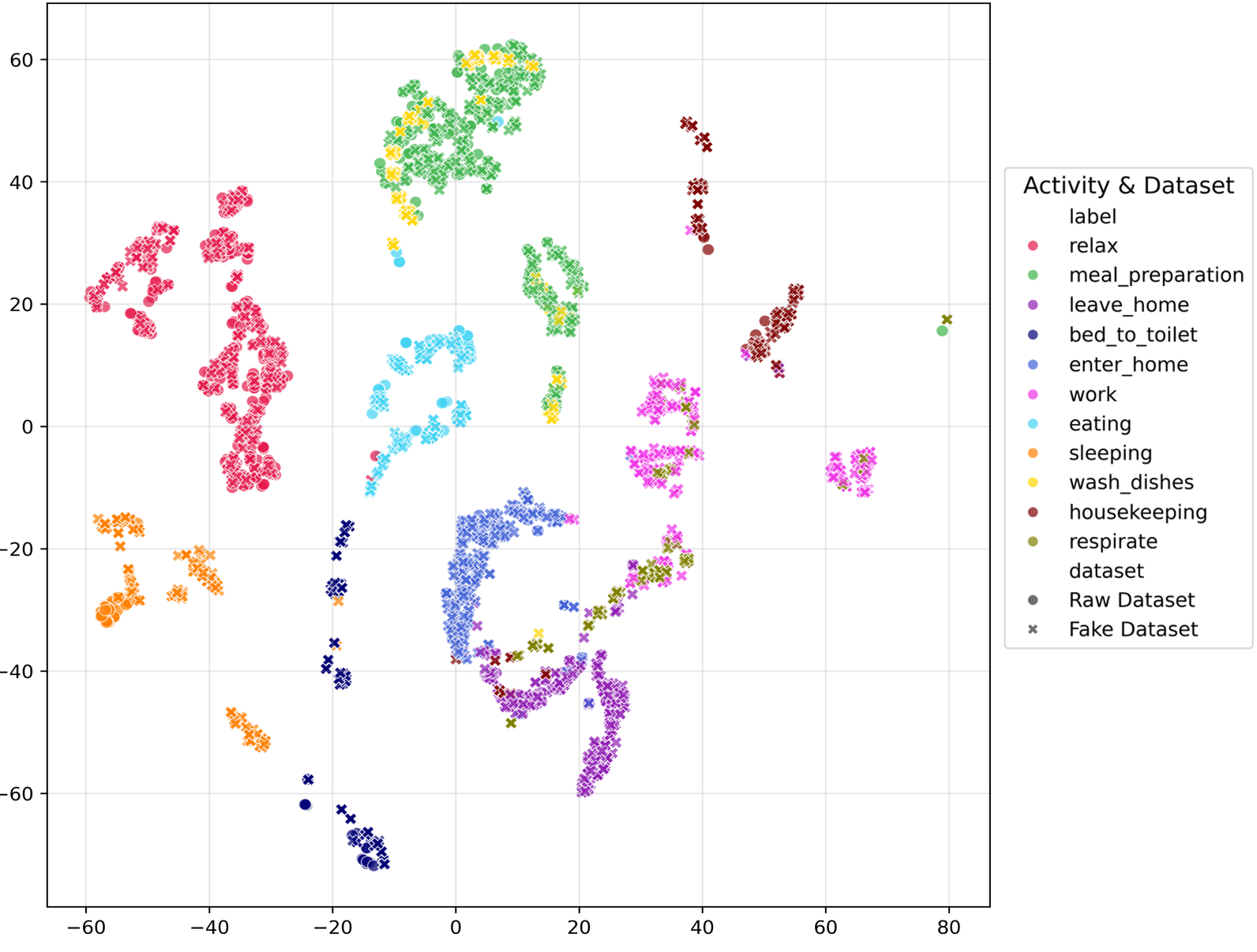} 
    \captionsetup{font=small}
    \caption{t-SNE visualizations of Bi-LSTM classifier (trained on real data) embeddings. The plots co-visualize real data ('O') and synthetic data ('X'), with activity classes consistently colored.}
    \label{fig:tsnes_bothset}
\end{wrapfigure}

\para{Behavioral Authenticity via Feature Space Visualization}
Figure~\ref{fig:tsnes_bothset} shows embeddings from a classifier trained on real data, with real samples and synthetic samples color-coded by activity class. The visualization reveals consistent co-clustering between real and synthetic instances of the same activity class, demonstrating that ADLGen effectively captures the underlying data structures. This strong distributional alignment renders synthetic sequences largely indistinguishable from real ones in discriminative feature spaces, supporting their potential for training robust ADL recognition models. Similar patterns are observed when testing the reverse scenario (training on synthetic, visualizing with real data). Meanwhile, the plot also highlights two important challenges: (1) "wash dishes" instances are difficult to distinguish from "meal preparation" activities, reflecting their semantic similarity and overlapping sensor patterns; and (2) the "respirate" class shows a somewhat chaotic distribution, attributable to its extremely limited representation in the original dataset (only 6 samples). Additional t-SNE visualizations have been included in the appendix.

\begin{table}[htbp]
\centering
\caption{\small Activity Recognition Performances.}
\label{tab:recognition_results_bilstm}
\setlength{\tabcolsep}{4pt}
\begin{tabular}{@{}llcc@{}}
\toprule
\textbf{Training Data} & \textbf{Test Data} & \textbf{MacroAcc (\%)} & \textbf{MacroF1 (\%)} \\
\midrule
\multicolumn{4}{@{}l}{\textit{Benchmark Performance}} \\
Real Aruba & Real Aruba & 79.78 & 79.73 \\
\midrule
\multicolumn{4}{@{}l}{\textit{Baselines (Trained on Synthetic variants)}} \\
Statistical Resampling & Real Aruba & 22.56 & 15.13 \\
LSTM Generator        & Real Aruba & 40.71 & 30.49 \\
VQVAE                 & Real Aruba & 49.58 & 39.67 \\
Transformer           & Real Aruba & 51.73 & 42.21 \\
cWGAN                 & Real Aruba & 55.08 & 45.12 \\
Seq2Res               & Real Aruba & 59.69 & 48.27 \\
\midrule
\multicolumn{4}{@{}l}{\textit{Our Method}} \\
ADLGen w/o Refine & Real Aruba & 65.27 & 50.77 \\
\rowcolor{LightGreen}ADLGen     & Real Aruba & 79.12 & 72.90 \\
\rowcolor{LightGreen}Real Aruba & ADLGen     & 79.89 & 73.12 \\
ADLGen            & ADLGen     & 98.74 & 98.72 \\
\bottomrule
\end{tabular}
\end{table}

\para{Downstream Activity Recognition.}
\label{sec:downstream_utility}

Using Esteban et al.'s synthetic framework \cite{esteban2017real}, we train a Bi-LSTM \cite{zhou2016bilstm} and report Macro-Acc/F1 in four settings: TSTR, TRTR, TRTS and TSTS (Train and Test settings, S = synthetic, R = real). As Table \ref{tab:recognition_results_bilstm} shows, the performance of TSTR matches TRTR nearly and exceeds all baseline generators, proving that ADLGen sequences encode the key activity cues needed for training. Real-trained models remain highly accurate on ADLGen data (TRTS), confirming its realism, while near-perfect TSTS scores indicate internal consistency, these findings position ADLGen as a valuable tool for developing robust ADL recognition systems while mitigating data scarcity and privacy concerns. Additional results regarding per-class metrics and the generalizability of synthetic data across different floorplans can be found in the Appendix for further reference.


\subsection{Scenario Simulation: Few-Shot Learning for Rare Activities}
\label{sec:simulation}
Real-world ADL datasets suffer from significant data volume constraints, with some activities being particularly scarce (e.g., only 33 samples for "Housekeeping" versus 2,900 for "Relax"). This scarcity makes it difficult to perform robust ADL downstream tasks, while collecting additional data is hindered by privacy concerns and high costs.
To demonstrate ADLGen's effectiveness in data-scarce environments, we simulate few-shot learning using extremely limited real samples for selected rare activities. ADLGen then generates synthetic sequences to augment these scarce training sets on the classification task, enabling effective model development without additional data collection.

\begin{table}[htbp]
\vspace{-10pt}
\centering
\caption{\small Few-Shot Classification of Rare Activities with ADLGen Data Augmentation. Performance is reported as F1-Score. Training with additional synthetic data is highlighted, best results are in \textbf{bold}.}
\label{tab:Scenario}
\small
\setlength{\tabcolsep}{3pt}
\renewcommand{\arraystretch}{0.8}
\begin{tabular}{lccccc}
\toprule
\multirow{2}{*}{\textbf{Training Data}} & \multicolumn{3}{c}{\textbf{F1-Score on Rare Activities$\uparrow$}} & \multirow{2}{*}{\textbf{MacroF1$\uparrow$}} & \multirow{2}{*}{\textbf{Improvement}} \\
 & \textbf{Housekeeping} & \textbf{Wash Dishes} & \textbf{Bed-Toilet} &  & \\
\midrule
\textbf{15 (Real Only)} & 0.55$\pm$0.01 & 0.75$\pm$0.02 & 0.98$\pm$0.01 & 0.76$\pm$0.02 & -- \\
\rowcolor{LightGreen}{\small \quad \textit{+100\% synthetic}} & 0.89$\pm$0.01 & 0.91$\pm$0.01 & 0.98$\pm$0.01 & 0.93$\pm$0.01 & +22.4\% \\
\rowcolor{LightGreen}{\small \quad \textit{+200\% synthetic}} & 0.93$\pm$0.02 & 0.96$\pm$0.01 & 1.00$\pm$0.00 & 0.96$\pm$0.01 & +26.3\% \\
\rowcolor{LightGreen}{\small \quad \textit{+400\% synthetic}} & \textbf{0.99$\pm$0.01} & \textbf{0.98$\pm$0.01} & \textbf{1.00$\pm$0.00} & \textbf{0.99$\pm$0.01} & \textbf{+30.2\%} \\
\midrule
\textbf{30 (Real Only)} & 0.90$\pm$0.01 & 0.88$\pm$0.01 & 0.98$\pm$0.01 & 0.92$\pm$0.01 & -- \\
\rowcolor{LightGreen}{\small \quad \textit{+100\% synthetic}} & 0.96$\pm$0.01 & 0.93$\pm$0.01 & 0.99$\pm$0.01 & 0.96$\pm$0.01 & +4.3\% \\
\rowcolor{LightGreen}{\small \quad \textit{+200\% synthetic}} & 0.98$\pm$0.01 & 0.96$\pm$0.01 & 1.00$\pm$0.00 & 0.98$\pm$0.01 & +6.5\% \\
\rowcolor{LightGreen}{\small \quad \textit{+400\% synthetic}} & \textbf{0.99$\pm$0.01} & \textbf{0.99$\pm$0.01} & \textbf{1.00$\pm$0.00} & \textbf{0.99$\pm$0.01} & \textbf{+8.6\%} \\
\bottomrule
\end{tabular}
\vspace{-8pt}
\end{table}
The results presented in Table \ref{tab:Scenario}, demonstrate that augmenting scarce real data with ADLGen-generated sequences yields substantial improvements in classification performance for rare activities. For instance, when starting with only 15 real samples per activity, the addition of 60 ADLGen-generated sequences boosts the overall MacroF1-score from 0.76 to an impressive 0.99. Similar significant gains are observed when starting with 30 real samples. This robust improvement, achieved even when ADLGen itself is trained on very limited data, suggests that our model effectively captures the underlying semantic and spatio-temporal characteristics defining each activity, rather than merely overfitting to the few available real instances. This capability is crucial for addressing the long-tail distribution of human behaviors in practical smart environment deployments.




\subsection{Ablation Study} 
\label{sec:ablation}

We conducted ablation experiments to assess each component's contribution to ADLGen performance.
As shown in Table \ref{tab:ablation}, sign-based tokenization proves most critical, with its removal causing severe degradation across all metrics, confirming its importance for encoding spatial-semantic relationships. Context-aware sampling ranks second, particularly for physical plausibility, highlighting its role in enforcing realistic spatial transitions. Temporal embedding ranks third, with its removal significantly affecting realism despite the limited variation in time-of-day/day-period within short-duration activities (<0.5 hour). Finally, while LLM refinement shows the smallest impact, this understates its value—even with already strong sequences, it still delivers substantial improvements, elevating semantic quality and achieving perfect physical plausibility. This highlights LLM refinement's capacity to enhance high-quality candidate sequences beyond what other components alone can achieve, demonstrating how each component addresses complementary aspects of ADL generation.

\begin{table}[htbp]
\centering
\caption{\small Ablation study on ADLGen components. Best scores are in \textbf{bold}.}
\label{tab:ablation}
\small
\setlength{\tabcolsep}{3pt}
\renewcommand{\arraystretch}{0.8}
\begin{tabular}{@{}lcccccc@{}}
\toprule
\multirow{2}{*}{\textbf{Model Variant}} & \multicolumn{2}{c}{\textbf{Realism (MMD$^2$$\downarrow$)}} & \multicolumn{2}{c}{\textbf{Validity Rate ($\uparrow$)}} & \multicolumn{2}{c}{\textbf{Semantic Score ($\uparrow$)}} \\
 & Value & $\Delta \%$ & Value & $\Delta$ & Value & $\Delta$ \\
\midrule
ADLGen (Full) & \textbf{0.0019$\pm$0.0009} & -- & \textbf{1.00} & -- & \textbf{4.67$\pm$0.22} & -- \\
\midrule
w/o Signed-based Tokenization   & 0.0880$\pm$0.0017 & 4630\%  & 0.49$\pm$0.04 & 0.51 & 1.95$\pm$0.24 & 2.72 \\
w/o Temporal Embedding          & 0.0059$\pm$0.0011 & 310.5\% & 0.76$\pm$0.05 & 0.24 & 3.49$\pm$0.25 & 1.18 \\
w/o Context-Aware Sampling      & 0.0053$\pm$0.0006 & 278.9\% & 0.46$\pm$0.07 & 0.54 & 2.84$\pm$0.15 & 1.83 \\
w/o LLM Refinement              & 0.0047$\pm$0.0005 & 247.4\% & 0.94$\pm$0.03 & 0.06 & 3.79$\pm$0.22 & 0.88 \\
\bottomrule
\end{tabular}%
\end{table}

\section{Conclusion}
We presented \textbf{ADLGen}, a novel generative framework specifically tailored for synthesizing realistic ADL sensor sequences in ambient assistive environments. ADLGen addresses the unique challenges—event-triggered sparsity, symbolic discreteness, spatial grounding, and activity-level semantics—through a decoder-only Transformer architecture with sign-based symbolic–temporal encoding and a context- and layout-aware sampling strategy. To ensure structural integrity and semantic plausibility, we further introduced a LLM-driven evaluation-and-refinement pipeline, which hierarchically verifies logical consistency, behavioral coherence, and temporal regularity, and automatically generates correction rules without requiring domain-specific tuning. We also proposed a comprehensive evaluation protocol encompassing statistical, diversity, semantic, spatial, and task-specific dimensions. Empirical results demonstrate that ADLGen not only generates high-quality and semantically rich sequences, but also significantly improves performance on downstream activity recognition tasks when used for data augmentation—offering a scalable, privacy-preserving alternative to conventional ADL data collection, and laying the foundation for broader adoption of generative models in ubiquitous computing and human-centered sensing applications.




\newpage
\appendix                 
\renewcommand\thesection{\Alph{section}}  
\setcounter{section}{0} 
\section*{Appendix}
\section{Experimental Setup and Implementation Details}
\label{app:exp_setup_impl_details}

\subsection{Dataset Details}
\label{app:dataset_details_in_setup}

\firstpara{CASAS Aruba Dataset.} The CASAS Aruba dataset~\cite{cook2012casas} was collected by Washington State University from November 2010 to June 2011. The smart home environment was equipped with 31 wireless binary sensors: 27 motion sensors (IDs prefixed with 'M'), 3 door sensors (IDs prefixed with 'D'), and 1 temperature sensor (ID prefixed with 'T'). The resident received regular visits from family members, which introduces realistic variability in the observed activity patterns.

\para{Sensor Layout.} The smart home's sensor layout informs our physical plausibility assessment. Motion sensors cover different functional areas (e.g., kitchen, living room, bedroom), while door sensors monitor entries/exits and essential cabinets. This spatial configuration creates inherent constraints that valid sensor sequences must satisfy. For example, sequential activations of distant motion sensors without intermediate sensor events would be physically implausible. Figure~\label{fig:aruba_layout} shows examples of floorplan layout (the layout of Aruba and Milan).

\begin{figure}[h]
    \centering
    \includegraphics[width=0.7\linewidth]{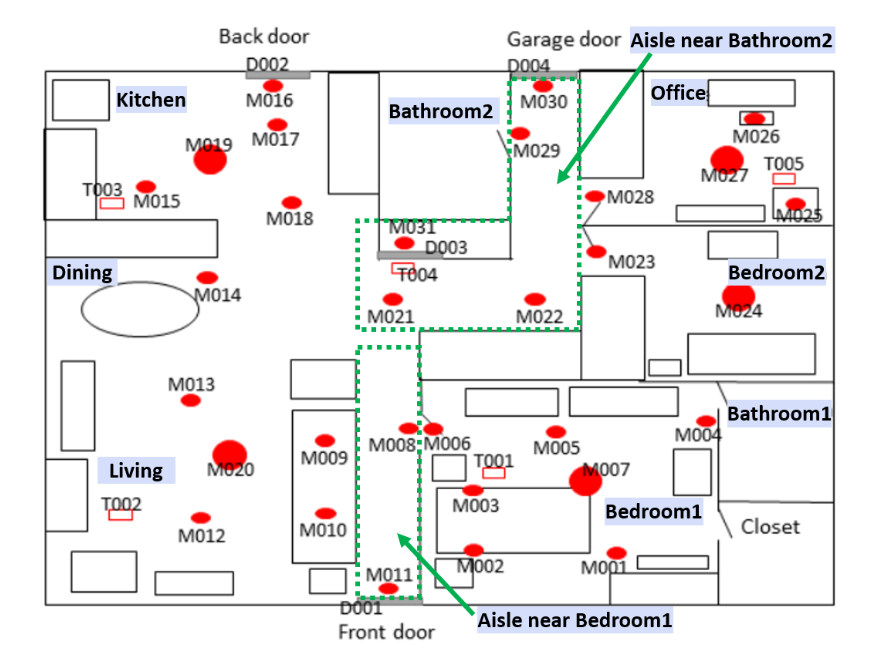}
    \caption{Sensor layout of Aruba floorplan}
    \label{fig:aruba_layout}
\end{figure}


\para{Activity Annotation Process.} Activities in the dataset were annotated using a combination of resident self-reporting and post-hoc verification by researchers. The annotation process involved marking the start and end times of specific activities, with periods outside these annotated windows labeled as "Other Activity." The annotation granularity varies across activities, with some having clear boundaries (e.g., "Enter home") and others being more diffuse (e.g., "Relax"). This annotation variability introduces challenges for traditional HAR approaches but provides our generative framework with rich contextual information about activity manifestation patterns.

\subsection{Data Preprocessing and Feature Engineering}
\label{app:preprocessing_details}

Our preprocessing pipeline involves several carefully designed steps to preserve temporal relationships while facilitating effective modeling. First, we perform temporal segmentation by extracting activity-labeled sequences through identifying all sensor events between activity start and end markers, resulting in sequences that vary in length from 2 to 87 events. For sensor status encoding, rather than using separate features for sensor ID and status, we create a unified encoding that integrates both—for instance, an "ON" status for sensor "M001" is encoded as "+M001" while an "OFF" status becomes "-M001". This approach increases vocabulary size but preserves the semantic relationship between sensor status.

For temporal feature extraction, we capture three key attributes for each sensor event: the hour of day (0-23) encoded using 4-bit cyclic encoding to preserve time continuity, the day of week (0-6) encoded using 3-bit cyclic encoding, and the event duration (time until next event) discretized into 8 logarithmic bins. When preparing sequences for training the generative model, we create input-target pairs using a sliding window approach with an input window of 10 events and a prediction target of the next event's sensor ID, status, and duration. Finally, we employ a stratified 5-fold cross-validation strategy to ensure representative activity distribution across all folds, with each sequence assigned entirely to either the training or test set to prevent information leakage.

\begin{algorithm}[h]
\caption{Detailed Preprocessing Pipeline}
\label{alg:preprocessing}
\begin{algorithmic}[1]
\Require $\mathcal{D}$: Raw sensor dataset with timestamps and activity labels
\Ensure $\mathcal{S}$: Processed sequences with temporal features

\Function{PreprocessDataset}{$\mathcal{D}$}
    \State $\mathcal{S} \gets \emptyset$ \Comment{Initialize processed sequences}
    \State $\mathcal{A} \gets \text{ExtractActivityBoundaries}(\mathcal{D})$
    
    \For{each activity instance $a \in \mathcal{A}$}
        \If{$a.\text{label} \neq \text{"Other Activity"}$}
            \State $E_a \gets \text{GetSensorEvents}(\mathcal{D}, a.\text{start}, a.\text{end})$
            \If{$|E_a| \geq 2$} \Comment{Ensure minimum sequence length}
                \State $S_a \gets \text{EncodeSequence}(E_a)$
                \State $S_a \gets \text{AddTemporalFeatures}(S_a)$
                \State $\mathcal{S} \gets \mathcal{S} \cup \{(S_a, a.\text{label})\}$
            \EndIf
        \EndIf
    \EndFor
    
    \State \Return $\mathcal{S}$
\EndFunction

\Function{EncodeSequence}{$E$}
    \State $S \gets \emptyset$
    \For{each event $e \in E$}
        \State $\text{encoded\_id} \gets e.\text{status} + e.\text{sensor\_id}$ \Comment{E.g., "+M001"}
        \State $S \gets S \cup \{\text{encoded\_id}\}$
    \EndFor
    \State \Return $S$
\EndFunction

\Function{AddTemporalFeatures}{$S$}
    \For{each event $s_i \in S$}
        \State $\text{hour} \gets \text{CyclicEncode}(s_i.\text{timestamp}.\text{hour}, 24)$
        \State $\text{day} \gets \text{CyclicEncode}(s_i.\text{timestamp}.\text{weekday}, 7)$
        \If{$i < |S|$}
            \State $\text{duration} \gets s_{i+1}.\text{timestamp} - s_i.\text{timestamp}$
            \State $\text{duration\_bin} \gets \text{LogBin}(\text{duration})$
        \Else
            \State $\text{duration\_bin} \gets 0$ \Comment{Default for last event}
        \EndIf
        \State $s_i.\text{temporal\_features} \gets (\text{hour}, \text{day}, \text{duration\_bin})$
    \EndFor
    \State \Return $S$
\EndFunction
\end{algorithmic}
\end{algorithm}

\subsection{Model Architecture}
We implement our framework using a decoder-only Transformer with $d = 384$ embedding dimension, $6$ attention heads, and $6$ decoder layers. Our sign-based representation uses vocabulary size $V = \text{num\_sensors} \times 2$ to encode both sensor identity and status. For temporal embeddings, we employ a shared codebook approach with learnable weights for hour of day, day of week, and temporal gaps. All embeddings are initialized with small random values ($\mathcal{N}(0, 0.02)$).

\subsection{Training and Generation Configuration}
Our model is trained using AdamW optimizer with base learning rate of $5 \times 10^{-4}$, weight decay $0.01$, and a OneCycleLR scheduler. We employ a batch size of $64$ and train for a maximum of $100$ epochs with early stopping based on validation loss. Generation uses nucleus sampling ($p = 0.95$), dynamic temperature (base $\tau = 0.7$), and repetition penalty ($\gamma = 1.1$) to balance sequence diversity with coherence. For semantic evaluation, a pre-trained language model converts sensor sequences to textual descriptions, with similarity computed via cosine distance in the embedding space. All models are trained on a single NVIDIA A100 GPU, with training times depending on dataset size.

\subsection{Formulation of Context-Aware and Spatially-Constrained Sampling}

We dynamically adjust the sampling temperature based on the evolving characteristics of the sequence being generated. 
At each decoding step $t$, the final sampling distribution is computed as:
\begin{align}
logits_{final}(s_{t}) &= \frac{logits(s_t)}{\tau_{t}}, ~\tau_{t} = \frac{\tau_{base}\cdot R(\boldsymbol {seq}^t_{sens})}{\alpha(L_{t-1}) \cdot \beta(D)}
\end{align}
$\tau_{t}$ is the proposed context-aware temperature, dynamically adjusted \textit{during generation} based on four contextual factors: 
\begin{enumerate} 
\item Base temperature $\tau_{base}$ controlling randomness; 
\item Length-aware scaling factor: 
\[\alpha(L_t) = \tau_{min} + \frac{L_{t} - L_{min}}{L_{max} - L_{min}} \cdot (\tau_{max} - \tau_{min})
\]
It encourages exploration in early decoding stages and coherence later (temperature $\tau_{max/min}$ are hyperparameters, and $L_{min}=3, L_{max}=100$);\\
\item diversity factor $\beta(D) \in \{1,1.2\}$ increases temperature when sensor ID diversity is low;\\
\item Repetition penalty 
\[
R(\boldsymbol {seq}^t_{sens})= {r_p \cdot (1.0 + 0.1 \cdot (n_{\boldsymbol {seq}^t_{sens}} - 1))}
\]

It penalizes overused sensors while allowing natural repetitions, here, $n_{\boldsymbol {seq}^t_{sens}}$ is the frequency of the current sensor id in generated sequence history, and $r_p$ is a tunable hyperparameter.\\
\end{enumerate}
This formulation allows temperature to evolve throughout the generation process, promoting diversity and flexibility in early stages and encouraging consistency and plausibility as the sequence unfolds.

\subsection{Generation Parameters for Minority Class Augmentation}
For minority class augmentation, we employ a sophisticated generation strategy that begins with class-conditional initialization, where the generation process is conditioned on the target activity class, using exemplar real sequences as prompts. We implement dynamic temperature control where the temperature is adjusted based on sequence diversity using the formula $\tau(s_{<t}) = \tau_b \cdot g(div(s_{<t}))$, where $\tau_b = 0.7$ is the base temperature, $div(s_{<t)})$ measures prefix diversity, and $g$ is a scaling function. Generated sequences undergo rigorous quality filtering to ensure they meet minimum quality thresholds, requiring a minimum semantic quality score of 3.5 out of 5, 100\% floorplan validity (no physically impossible sensor activations), and a minimum diversity score of 0.4.

\section{Symbolic-Temporal Decoupling for ADL Representation}
\label{sec:discussion_decoupling}

\subsection{Motivation and Limitations of Interleaved Approaches}
\label{subsec:motivation_decoupling}

\begin{figure}
    \centering
    \includegraphics[width=.6\linewidth]{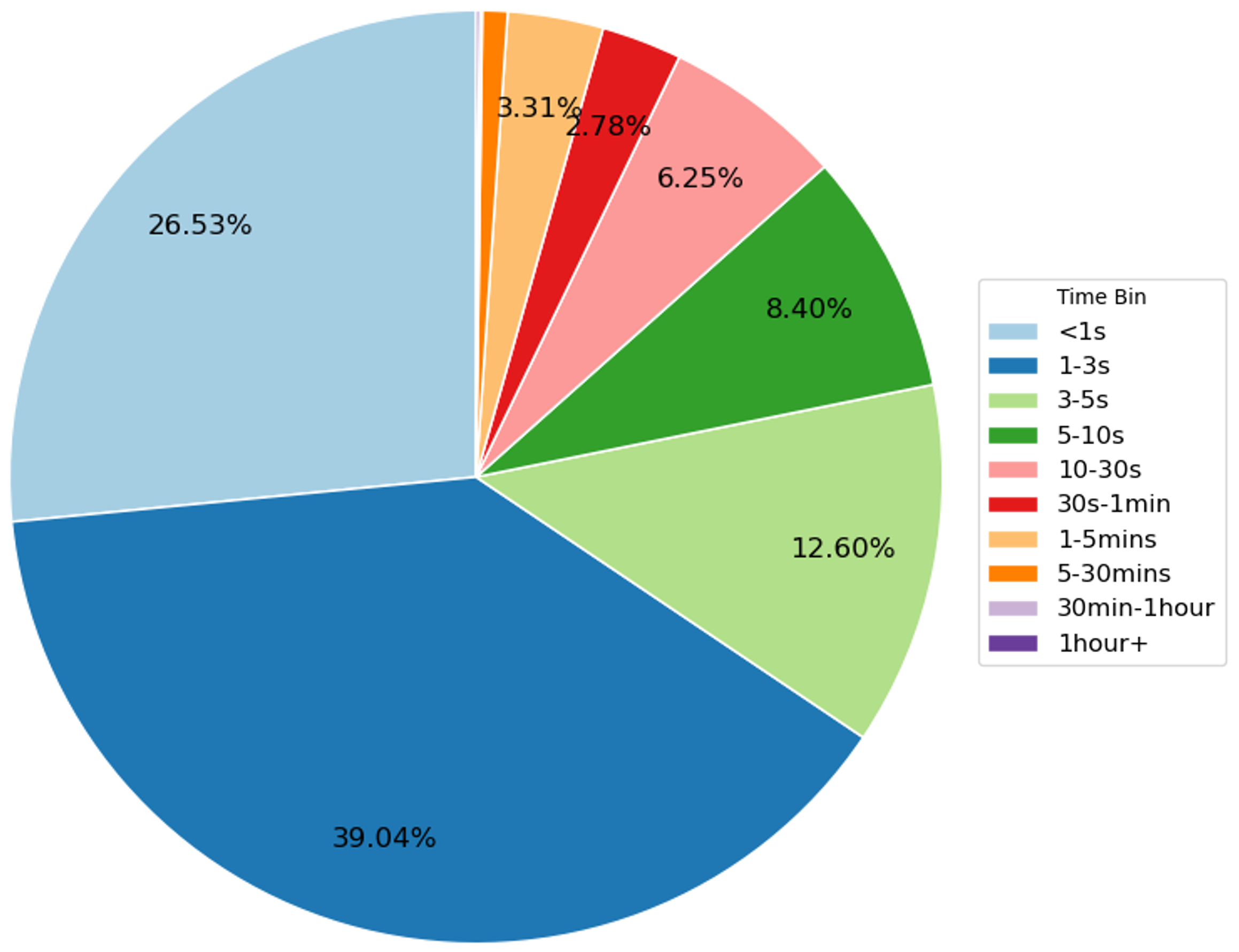}
    \caption{\small Distribution of sensor event intervals.}
    \label{fig:time_delta}
\end{figure}

Our proposed symbolic-temporal decoupling approach separates sensor event sequences from their temporal markers to create more effective representations for transformer-based ADL modeling. This section provides the theoretical motivation and analyzes the benefits of this design choice.

ADL event streams contain two distinct information types: symbolic data (which sensor fired and its status) and temporal data (when the event occurred). Standard sequence modeling approaches typically flatten this structured information into a single interleaved sequence (e.g., \texttt{[SensorID, SensorStatus, Time, SensorID, ...]}), embedded using token and positional encodings.

While transformers can theoretically learn from any sequence structure, we identify several fundamental limitations in the ADL context:

\begin{enumerate}[topsep=0pt,itemsep=0.3em,parsep=0pt,leftmargin=*]
    \item \textbf{Information Asymmetry:} Sensor (ID+status) patterns typically carry the most discriminative information for ADL identification, while temporal features often remain static throughout an activity (e.g., an entire "bed to toilet" sequence occurring within the same hour). In an interleaved representation, these repeated discrete temporal tokens contribute minimal new information while potentially diluting attention to critical sensor patterns.
    
    \item \textbf{Positional Encoding Mismatch:} For sensor events, sequence position carries crucial information about activity structure (similar to word order in language). However, temporal tokens derive meaning from their absolute value rather than their position in sequence—"10 AM" carries the same semantic meaning regardless of whether it appears as the 1st or 5th token. Standard positional encoding thus provides valuable information for sensor tokens but less meaningful for temporal tokens.
    
    \item \textbf{Representational Interference:} Embedding fundamentally different data types—discrete symbolic sensor IDs and continuous/cyclical temporal values—into a shared representation space forces suboptimal compromises. The optimal embedding space for sensor logic likely differs from that for temporal progression, leading to representational trade-offs.
    
    \item \textbf{Conflated Temporal Dynamics:} Temporal patterns (e.g., durations, intervals, time-of-day effects) follow different statistical properties than the logical, grammar-like sequences of sensor events that define an activity. A single attention mechanism operating on a mixed stream struggles to disentangle and model these distinct dynamics effectively.

    \item \textbf{Temporal Granularity and Sparsity:} While fine-grained timestamps might seem preferable to hour-of-day discretization, they introduce significant challenges for transformer-based modeling. With most ADL sensor events occurring within 5-second intervals (as shown in Figure~\ref{fig:time_delta}, over 26\%$<$1s and 78\%$<$5s), precise timestamps would create an extremely sparse, long-tailed token distribution (across 24 hours × 60 minutes × 60 seconds possible values). This sparsity presents fundamental statistical learning difficulties, as many timestamp tokens would appear too infrequently to learn meaningful representations. Additionally, the most precise temporal patterns often reflect sensor sampling artifacts rather than meaningful activity characteristics, potentially harming generalization and diversity.
\end{enumerate}

\subsection{Theoretical Advantages of Symbolic-Temporal Decoupling}
\label{subsec:rationale_decoupling}

Our decoupling approach processes symbolic sensor sequences ($\boldsymbol{seq}_{\text{sens}}$) and temporal sequences ($\boldsymbol{seq}_{\text{time}}$) separately before fusion:
\[
E_{\text{total}} = E_{\text{sens}}(\boldsymbol{seq}_{\text{sens}}) + \omega_p \cdot PE(\boldsymbol{seq}_{\text{sens}}) + \omega_t \cdot E_{\text{temp}}(\boldsymbol{seq}_{\text{time}})
\]

This design offers several theoretical advantages:

\begin{enumerate}[topsep=0pt,itemsep=0.3em,parsep=0pt,leftmargin=*]
    \item \textbf{Specialized Representations:} Decoupling enables dedicated pathways for modeling sensor sequences and temporal patterns. The symbolic pathway ($E_{\text{sens}} + PE$) applies positional encoding to sensor tokens where sequence order is semantically meaningful, while the temporal pathway ($E_{\text{temp}}$) can encode time based on its intrinsic properties (e.g., cyclical encodings for hour-of-day) rather than its sequence position.
    
    \item \textbf{Adaptive Information Weighting:} Temporal information acts as a contextual modulator through $\omega_t$ rather than competing directly with sensor tokens for attention. The model can learn to dynamically adjust the influence of temporal context based on its relevance to each prediction step.
    
    \item \textbf{Enhanced Sequence Modeling:} By maintaining the contiguity of sensor events, the model can better capture the logical "grammar" of activities without temporal tokens fragmenting these patterns. This aligns with findings in multimodal learning, where modality-specific feature extraction before fusion often outperforms early modality mixing.
\end{enumerate}

Our preliminary experiments strongly support these theoretical advantages. We observed that interleaved approaches using only hour-of-day temporal information resulted in significant convergence difficulties and substantially higher training losses compared to our decoupled architecture. These findings, while qualitative, provide compelling evidence that symbolic-temporal decoupling is essential for effective modeling of ADL sequences, particularly when using coarse-grained temporal features like hour-of-day that remain static across many consecutive sensor events.

\section{Sensor Tokenization Strategies}

\subsection{Comparison of Sign-Based and Composite Vocabulary Tokenization}

In this section, we analyze two primary approaches for tokenizing sensor status in activity recognition models: (1) our sign-based approach where sensor status are represented by the sign of the token value, and (2) a composite vocabulary approach that assigns unique vocabulary entries to each sensor-status combination.

\subsubsection{Sign-Based Tokenization}

Our current implementation uses a sign-based tokenization scheme where:

\begin{itemize}
    \item Sensor identities are represented by the absolute value of the token
    \item Sensor status are encoded by the sign (positive for ON/OPEN, negative for OFF/CLOSED)
    \item The embedding process involves two separate embedding tables (sensor and status)
    \item Each embedding utilizes $d_{\text{model}}/2$ dimensions, which are later concatenated and projected
\end{itemize}

This approach offers several advantages:

\begin{enumerate}[topsep=0pt,itemsep=0.3em,parsep=0pt,leftmargin=*]
    \item \textbf{Parameter Efficiency:} With approximately 40 sensors, our approach requires $O(|S|)$ parameters in the embedding tables, where $|S|$ is the number of sensors and $|V|$ is the number of possible status values.
    
    \item \textbf{Status Generalization:} By sharing status embeddings across all sensors, the model better generalizes status information, particularly valuable for sensors with imbalanced sensor status distributions (In a sensor stream-like ADL dataset, activity sequences are manually labeled from beginning to end, which results in unpaired sensor status counts in most sequences.).
    
    \item \textbf{Multi-Status Support:} The architecture naturally extends beyond binary status values, accommodating sensors with multiple possible status without vocabulary explosion.
    
    \item \textbf{Information Disentanglement:} Separating sensor identity from status information creates more interpretable embeddings and potentially enables better transfer learning.
\end{enumerate}

\subsubsection{Composite Vocabulary Approach}

An alternative approach assigns unique tokens for each sensor-status combination (e.g., M001\_ON, M001\_OFF):

\begin{itemize}
    \item Each sensor-status pair receives a unique token ID
    \item A single embedding table maps directly to the full embedding dimension
    \item Generation requires only a single prediction step rather than predicting sensor and status separately
\end{itemize}

This approach may offer advantages in certain scenarios:

\begin{enumerate}[topsep=0pt,itemsep=0.3em,parsep=0pt,leftmargin=*]
    \item \textbf{Direct Probability Modeling:} With well-balanced datasets, direct modeling of sensor-status transitions can potentially yield more coherent generated sequences.
    
    \item \textbf{Computational Simplicity:} The approach requires a single embedding lookup rather than multiple embedding operations and projections.
    
    \item \textbf{Status-Specific Patterns:} Treating each Status as unique might better capture tight correlations between specific sensor Status.
\end{enumerate}

\subsubsection{Superiority Over Naive (Flatten) Tokenization}

Both approaches significantly outperform the flatten method of representing sensors and their Status as separate tokens in a sequence (e.g., [M001, ON, M001, OFF]):

\begin{itemize}
    \item \textbf{Sequence Length Efficiency:} Both our approaches halve the sequence length compared to the flatten approach, enabling longer effective context windows and more efficient attention mechanisms.
    
    \item \textbf{Temporal Coherence:} Pairing sensors with their Status preserves the temporal integrity of the data, preventing the model from producing invalid sequences where Status are disconnected from their sensors.
    
    \item \textbf{Reduced Ambiguity:} In the flatten approach, the model must learn which Status tokens apply to which sensor tokens, introducing unnecessary ambiguity and complexity.
    
    \item \textbf{Improved Long-Range Dependencies:} Shorter sequences allow the attention mechanism to more effectively capture long-range dependencies between sensor events.
    
    \item \textbf{Generation Quality:} Both approaches prevent the generation of invalid intermediate status (such as consecutive ON tokens without intervening sensor tokens, [M001, ON, ON, M001]).
\end{itemize}

\subsection{Trade-offs and Performance Implications}

Our analysis reveals important trade-offs between these approaches:

\begin{align}
\text{Vocabulary Size:} &
\begin{cases}
O(|S| + |V|) & \text{Sign-based approach} \\
O(|S| \times |V|) & \text{Composite vocabulary approach} \\
O(|S| + |V|) & \text{Flatten tokens approach}
\end{cases}
\end{align}

\begin{align}
\text{Sequence Length:} &
\begin{cases}
n & \text{Sign-based approach} \\
n & \text{Composite vocabulary approach} \\
2n & \text{Naive flatten tokens approach}
\end{cases}
\end{align}

While composite vocabulary tokenization theoretically enables more direct modeling of status transitions, our sign-based approach offers superior performance in several key scenarios:

\begin{itemize}
    \item \textbf{Handling Data Sparsity:} When certain sensor-status combinations are rare, our approach leverages shared status representations, whereas the composite vocabulary approach may struggle with rare combinations.
    
    \item \textbf{Generalization to Unseen Patterns:} By factorizing sensor identity and status, our model can better generalize to combinations not seen during training.
    
    \item \textbf{Scaling to Additional Sensors:} Adding new sensors requires minimal adjustment to our embedding architecture.
    
    \item \textbf{Supporting Non-Binary Status:} Our approach naturally extends to sensors with continuous or multi-valued status.
\end{itemize}

However, it should be noted that with high-quality datasets exhibiting balanced sensor-status distributions and primarily binary sensors, the composite vocabulary approach could potentially produce more coherent generated sequences due to its direct modeling of status transitions. This advantage becomes particularly relevant in controlled environments with consistent sensor activation patterns and abundant training data.

\begin{table}[h]
\centering
\caption{Comparison of Sensor Tokenization Methods (w/o temporal in the sequence)}
\begin{tabular}{lccc}
\toprule
\textbf{Metric} & \textbf{Sign-Based} & \textbf{Composite} & \textbf{Flatten} \\
\midrule
Vocabulary Size (40 sensors) & $\sim$40 tokens & $\sim$80 tokens & $\sim$42 tokens \\
Parameters & $O(|S|)$ & $O(|S| \times |V|)$ & $O(|S| + |V|)$ \\
Computation & Two embed. + proj. & Single lookup & Two lookups \\
Sequence Length & $n$ & $n$ & $2n$ \\
\midrule
Sequence Coherence & Good & Very Good & Poor \\
Rare Sensor Handling & Strong & Weak & Moderate \\
Generalization & Very Good & Limited & Good \\
Multi-Status Support & Excellent & Poor & Good \\
\midrule
Perplexity (balanced data) & Moderate & Lower & Higher \\
Generation Diversity & Higher & Lower & Higher \\
Training Speed & Slower & Faster & Slowest \\
Scaling to More Sensors & Good & Degrades & Moderate \\
\midrule
Valid Sequence Rate & 91-96\% & 90-94\% & 56-59\% \\
Semantic Score & 3.55-3.76 & 3.21-3.48 & 2.23-2.62 \\
\bottomrule
\end{tabular}
\label{table:tokenization_comparison}
\end{table}

Our sign-based tokenization approach represents a more robust and generalizable solution for human activity recognition, particularly in real-world settings with varied sensor distributions and potential data sparsity issues. The factorization of sensor identity and status creates more interpretable embeddings and enables better handling of the inherent diversity in sensor data. While composite vocabulary tokenization may offer advantages in idealized conditions with abundant training data, our approach better addresses the challenges of real-world activity recognition scenarios. Both approaches substantially outperform the flatten method of representing sensors and status as separate tokens, offering significant advantages in sequence length efficiency, temporal coherence, and generation quality.

\section{Detailed Implementation of Evaluation Metrics}
\label{app:metric_implementation}

\subsection{Diversity Score Calculation}
\label{app:diversity_calc}

Our diversity score quantifies sequence variation across multiple dimensions, providing a more nuanced assessment than simple entropy measures. The calculation begins by identifying four key components for analysis: sensor ID distribution (which sensors are activated), status transitions (patterns of ON/OFF activations), temporal distribution (when activities occur), and sequential patterns (order of sensor activations).

For each component $k$, we calculate Shannon entropy using the formula:
\begin{equation}
H_k = -\sum_{i=1}^{n} p_i \log_2 p_i
\end{equation}
where $p_i$ represents the probability of occurrence for the $i$-th element in component $k$. 

We then determine the theoretical maximum entropy for each component as:
\begin{equation}
H_{max_k} = \log_2(n_k)
\end{equation}
where $n_k$ is the number of unique elements in component $k$. 

Next, we calculate normalized entropy for each component:
\begin{equation}
H'_k = \begin{cases}
\frac{H_k}{H_{max_k}} & \text{if } H_{max_k} > 0 \\
0 & \text{otherwise}
\end{cases}
\end{equation}

To compute the final diversity score, we create the set $S_{+}$ containing only those normalized entropies that are strictly greater than 0:
\begin{equation}
S_{+} = \{H'_k \mid k \in C \text{ and } H'_k > 0\}
\end{equation}

The diversity score is then calculated as the geometric mean of values in $S_{+}$:
\begin{equation}
\text{Diversity Score} = \left( \prod_{H'_k \in S_{+}} H'_k \right)^{1/|S_{+}|}
\end{equation}

This geometric mean provides a balanced assessment that penalizes poor diversity in any single component, encouraging well-rounded diversity across all aspects of the generated sequences.

\subsection{Unbiased estimation of Square MMD}

The Maximum Mean Discrepancy (MMD) is a non-parametric metric used to measure the distance between two probability distributions $P$ and $Q$ based on samples $X = \{x_1, \ldots, x_m\} \sim P$ and $Y = \{y_1, \ldots, y_n\} \sim Q$. It is widely used to evaluate the similarity between real and generated data in generative modeling.

\textbf{Unbiased estimator (without diagonal terms):}
\[
\text{MMD}^2(X, Y) = \frac{1}{m(m-1)} \sum_{i \neq j} k(x_i, x_j)
+ \frac{1}{n(n-1)} \sum_{i \neq j} k(y_i, y_j)
- \frac{2}{mn} \sum_{i=1}^m \sum_{j=1}^n k(x_i, y_j)
\]

Here, $k(\cdot, \cdot)$ is a positive-definite kernel function (e.g., Gaussian kernel), which measures the similarity between samples. The first two terms compute the average similarity within each sample set (excluding self-similarity, i.e., $i \neq j$), while the last term computes the average similarity between the two sets. Excluding diagonal terms ensures the estimator is unbiased.

\textbf{Biased estimator (with diagonal terms):}
\[
\text{MMD}^2(X, Y) = \frac{1}{m^2} \sum_{i=1}^m \sum_{j=1}^m k(x_i, x_j)
+ \frac{1}{n^2} \sum_{i=1}^n \sum_{j=1}^n k(y_i, y_j)
- \frac{2}{mn} \sum_{i=1}^m \sum_{j=1}^n k(x_i, y_j)
\]

In the biased estimator, the diagonal terms ($i=j$) are included in the within-set sums. This version is slightly easier to compute and commonly used in practice, but it introduces a small bias, especially for small sample sizes.

\textbf{Interpretation:}  
A \textbf{lower} MMD value indicates that the two distributions are \textbf{more similar} according to the chosen kernel. The choice of kernel and its parameters (e.g., bandwidth for the Gaussian kernel) can significantly affect the sensitivity of the MMD metric.

\subsection{Semantic Quality Metrics}
The semantic quality of a generated sequence $s$ for activity $a$ is formally evaluated using a weighted combination of three quality dimensions:
\begin{align}
  \text{SemanticQuality}(s, a) = \alpha_F \phi_F + \alpha_S \phi_S + \alpha_T \phi_T
\end{align}

where $\phi_F$, $\phi_S$, and $\phi_T$ represent functional, sequential, and temporal quality scores, with corresponding weights $\alpha_F$, $\alpha_S$, and $\alpha_T$. The functional dimension evaluates whether the sequence contains the essential sensor activations characteristic of the activity (e.g., bathroom sensors for bathing). The sequential dimension assesses the logical progression of sensor events (e.g., opening a cabinet before removing an item). The temporal dimension examines whether the timing patterns align with typical activity durations and time-of-day distributions.

\subsection{Intra-set (activity classes) similarity Metrics}
Given a set of $N$ sequences in an activity class, we first compute a pairwise similarity matrix $S \in \mathbb{R}^{N \times N}$, where each entry $S_{ij}$ represents the similarity (e.g., cosine similarity) between sequence $i$ and sequence $j$. The diagonal entries $S_{ii}$ represent self-similarity and are excluded from metric calculations.

\textbf{Formulation:}
Let $S^* = \{ S_{ij} \mid i \neq j \}$ denote the set of all off-diagonal similarity values.\\
\textbf{Mean similarity:}
\[\mu = \frac{1}{N(N-1)} \sum_{i=1}^N \sum_{\substack{j=1 \\ j \neq i}}^N S_{ij}\]
\textbf{Median similarity:}
\[\text{Median} = \mathrm{median}(S^*)\]
\textbf{Minimum similarity:}
\[\text{Min} = \min(S^*)\]
\textbf{Maximum similarity:}
\[\text{Max} = \max(S^*)\]
\textbf{Standard deviation:}
\[\sigma = \sqrt{ \frac{1}{N(N-1)} \sum_{i=1}^N \sum_{\substack{j=1 \\ j \neq i}}^N (S_{ij} - \mu)^2 }\]
\textbf{Percentiles:}
\[P_{25} = \text{25th percentile of } S^* \qquad P_{75} = \text{75th percentile of } S^*\]

\textbf{Interpretation:}  
These metrics summarize the distribution of pairwise similarities within a set of sequences, excluding self-similarity. The mean and median provide measures of central tendency, while the minimum and maximum indicate the range of similarities. The standard deviation quantifies the spread, and the percentiles offer additional insight into the distribution. Higher mean or median values indicate that the sequences in the set are more similar to each other, while lower values suggest greater diversity.

\newpage
\section{LLM Evaluation and Refinement Details}
\label{app:llm_eval_refine_details}

\subsection{LLM-based Evaluation Algorithm}
Pseudo code in Algorithm~\ref{alg:semantic_eval_full} details our semantic evaluation and refinement process.

\begin{algorithm}[H]
\small
\caption{LLM-based Semantic Evaluation and Refinement}
\label{alg:semantic_eval_full}
\begin{algorithmic}[1]
\Require $\mathcal{S}$: Generated sensor sequences, $\mathcal{A}$: Activity labels, $\mathcal{E}$: Environmental context
\Ensure Refined sequences and evaluation metrics

\Function{EvaluateAndRefine}{$\mathcal{S}$, $\mathcal{A}$, $\mathcal{E}$}
    \State $\mathcal{R}_{results} \gets \emptyset$ \Comment{Results container}
    \State $\mathcal{S}_{refined} \gets \emptyset$ \Comment{Refined sequences container}
    
    \For{$(s, a)$ in $(\mathcal{S}, \mathcal{A})$}
        \State $text \gets \text{Sequence2Text}(s, \mathcal{E})$ \Comment{Convert to text description}
        
        \State $\phi_P \gets \text{EvaluatePhysicalPlausibility}(text, \mathcal{E})$ \Comment{Spatial constraints}
        \State $\phi_B \gets \text{EvaluateBehavioralCoherence}(text, a)$ \Comment{Activity consistency}
        \State $\phi_T \gets \text{EvaluateTemporalConsistency}(text, a)$ \Comment{Timing patterns}
        
        \State $issues \gets \text{IdentifyConstraintViolations}(text, \phi_P, \phi_B, \phi_T)$
        \State $rules \gets \text{GenerateRefinementRules}(issues, \mathcal{E})$ \Comment{Auto-generate rules}
        
        \State $s_{refined} \gets s$ \Comment{Start with original sequence}
        \For{$rule$ in $rules$}
            \State $s_{refined} \gets \text{ApplyRule}(s_{refined}, rule)$ \Comment{Programmatic refinement}
        \EndFor
        
        \State $\mathcal{S}_{refined} \gets \mathcal{S}_{refined} \cup \{s_{refined}\}$
        \State $\mathcal{R}_{results} \gets \mathcal{R}_{results} \cup \{(\phi_P, \phi_B, \phi_T, rules)\}$
    \EndFor
    
    \State \Return $\mathcal{S}_{refined}, \mathcal{R}_{results}$
\EndFunction
\end{algorithmic}
\end{algorithm}

\subsection{Tiered Semantic Evaluation Approach}
Our tiered evaluation methodology first establishes foundational viability by scrutinizing basic physical plausibility and logical sensor status progression. Subsequently, sequences that pass these initial checks undergo a more in-depth assessment of their behavioral coherence, temporal consistency, and detailed semantic alignment with the specific ADL context.

\subsection{Rule Generation and Application Details}
For identified violations, the LLM automatically generates executable refinement rules. These rules are denoted as:
\begin{align}
    \mathcal{R}=\{r_k=\text{LLM}(\boldsymbol{Seqs}, \Phi)\,|\,k=1,\dots,K\}
\end{align}
where $\Phi$ provides the LLM context during semantic assessment. Each rule in $\mathcal{R}$ is designed to address a specific violation type of error within $\boldsymbol{Seqs}=\{\boldsymbol{seq_1}, \boldsymbol{seq_2},....\}$, resulting in a corrected sequence set $\boldsymbol{Seqs'}$.
To apply the rules, we suggest three options:
\begin{enumerate}[topsep=0pt,itemsep=0.3em,parsep=0pt,leftmargin=*]
    \item Utilize the prompt \ref{refinement_prompt} to enable the LLM to automatically modify the violations (used in our experiment).
    \item Employ the Python script provided by LLM to handle the sequence set.
    \item Manually write a rule-based script following the rules to make the corrections. 
\end{enumerate}
The rule types include:
insertion rules adding missing logical preconditions (e.g., incorporating door sensor activation), deletion rules eliminating physically impossible events, and reordering rules rectifying temporal inconsistencies. Leveraging the LLM’s instruction-following capability, this refinement process is fully automated and generalizes across different environments without requiring task-specific rules or domain expertise.

\newpage
\subsection{LLM Evaluation Prompt for Semantic Scoring}
In our study, the LLM from Google Gemini family: Gemimi 2.5 Pro Preview, which accommodates more than 1 million tokens of context, enabling the input of our lengthy activity sequences. To replicate our assessment, we suggest testing the prompt with the API of Google Gemini 2.5 Pro Preview 05-06 model and configuring structured output.\\
\\
\begin{tcolorbox}[title=LLM Evaluation Prompt, fonttitle=\bfseries, colback=gray!5!white, colframe=black, listing only, listing options={%
    basicstyle=\ttfamily\small,
    breaklines=true,
    language=Markdown,
    showstringspaces=false,
    escapeinside=||
}]
\label{scoring_prompt}
**System Prompt:**\\
You are an AI assistant specialized in evaluating the semantic quality of sensor event sequences representing human Activities of Daily Living (ADLs). Use world knowledge of human behaviors and spatial reasoning. **Focus on the semantic consistency of the provided sequence segment and its alignment with the activity label, recognizing sequences may be truncated or contain real-world noise.** Respond formally and directly.\\
\\
**Task:**\\
With input **Sample ID: [Activity label] sensor sequence**, analyze EACH provided sensor sequence and its associated activity label with a given floorplan map based on the criteria below. You will receive MULTIPLE samples in the input. Calculate an integer score (1–5) for each.\\
\\
**Output Format:**\\
For EACH sample analyzed, output its Sample ID and calculated score on a **new line**, formatted strictly as a comma-separated value (CSV) row: `Activity,SampleID,Score`.\\
**Output ONLY the CSV lines, one per sample.** Do NOT include a header row. Do NOT include explanations, introductory text, or any other characters.\\
\\
**Evaluation Principles:**\\
1.  **Sequence Segment Focus:** Evaluate the quality and consistency of the **provided segment**. Do not penalize solely for incompleteness if the segment is valid.\\
2.  **Sensor Interpretation:** `ON`=person detected; `OFF`=person not detected (left/still); `Door OPEN/CLOSE`=literal.\\
3.  **Redundancy Tolerance:** Limited consecutive identical reports are acceptable; focus on status changes and overall logic.\\
\\
**Scoring Criteria (Apply Sequentially):**\\
**A. Foundational Plausibility Checks (Determines if Score is 1–2 vs 3–5):**\\
  -**Status Transition Logic:** 'ON' must alternate with 'OFF' for the same sensor ID.\\
  -**Spatial Coherence:** No implausible physical jumps.\\
  -**Logical Reasonableness:** Overall sequence must be behaviorally coherent.\\
  -**→ If violated, assign Score 1–2 and stop.**\\
\\
**B. Semantic Quality \& Behavior Alignment (Scores 3–5 only if A is passed):**\\
-**Score 5:** High internal consistency and strong alignment with the labeled activity.\\
-**Score 4:** Valid segment with slight deviations or partial alignment.\\
-**Score 3:** Weak alignment, low sensor diversity, or internally inconsistent but not illogical.\\
\end{tcolorbox}

\newpage
\subsection{LLM Refinement Prompt for Rule Generation and Correction}
\begin{tcolorbox}[title=LLM Refinement Prompt, fonttitle=\bfseries, colback=gray!5!white, colframe=black, listing only, listing options={%
    basicstyle=\ttfamily\small,
    breaklines=true,
    language=Markdown,
    showstringspaces=false,
    escapeinside=||
}]
\label{refinement_prompt}
**System Prompt:**\\
You are an AI assistant tasked with refining sensor event sequences representing human Activities of Daily Living (ADLs). You have already performed a detailed semantic evaluation of each sequence using a hierarchical quality function. You now must generate and apply executable rules to correct semantic violations.\\
\\
**Task:**\\
Using the evaluation context already provided (including Quality scores and dimension-specific insights), refine all sequences in the dataset that received scores below 5. For each low-quality sequence, generate one or more correction rules that target the identified issues, then apply these rules to produce a refined version of the sequence.\\
\\
**Refinement Strategy Based on Score:**\\
- **Score 1–2:** Focus on fixing fundamental physical and logical violations, such as invalid sensor transitions, impossible spatial paths, or major inconsistencies.\\
- **Score 3–4:** Focus on correcting higher-level semantic issues, such as incomplete or misaligned behavior patterns, implausible sensor diversity, or incorrect temporal structure.\\

**Output Format:**\\
For each sequence, output two blocks:\\
\\
1. **Rule Set Block (one or more lines):**\\
    `Activity,SampleID,RuleType,Target,Explanation`\\
    -   $\text{RuleType} \in {\text{INSERT, DELETE, REORDER}}$\\
    - Target specifies the relevant sensor ID(s) and timestamp(s)\\
\\
2. **Refined Sequence Block (1 line):**\\
    `Activity,SampleID,[SensorID1,Status1,Time1];[SensorID2,Status2,Time2];...`\\
\\
- Separate rules and refined sequence with a blank line.\\
- Output strictly in the specified format. No extra commentary or headers.\\
\\
**Guidelines:**\\
- Use your world knowledge and prior Scoring Criteria:

\hspace{1cm} $A(\text{Foundational Plausibility})$ and $B(\text{Semantic Quality \& Behavior Alignment})$\\
- Do not hallucinate behaviors unrelated to the original activity label.\\
- Minimize overcorrection—retain plausible noise and natural variation.\\
- Ensure applied rules resolve all major flagged violations per their score class.\\
- Process and refine every input sequence in the provided set.\\
\\
**[Input will include evaluated sequences with Activity Label, Sample ID, original sequence, and evaluation score]**\\
\end{tcolorbox}
\newpage

\subsection{Example of Reasoning Context during LLM Evaluation}
\begin{tcolorbox}[
    title=LLM Evaluation Reasoning Context, 
    fonttitle=\bfseries, 
    colback=gray!5!white, 
    colframe=black
]
\label{llm_scoring_response} 

\paragraph{Activity:}\textbf{Eating}
\newline
\textbf{Core essence:} Person in Dining area. May involve movement from kitchen or living room.

\smallskip 

\noindent\textbf{Sensors involved:} M014 (``In Dining area''), M020 (``In the center of the Living room'').

\paragraph{Sample 1} 
\begin{quote}
\itshape 
``The motion sensor In Dining area reported a 'ON' status during morning.'', 
``The motion sensor In Dining area reported a 'OFF' status during morning.'', 
``The motion sensor In Dining area reported a 'ON' status during morning.'' \ldots{} 
``The motion sensor In the center of the Living room reported a 'OFF' status during morning.''
\end{quote}

\begin{itemize}[leftmargin=*, topsep=3pt, itemsep=3pt] 
    \item \textbf{Check A:}
        \begin{itemize}[leftmargin=*, itemsep=1pt]
            \item \textbf{Status Transition:}
                \begin{itemize}[leftmargin=*, itemsep=0pt]
                    \item Sensor (``In Dining area''): ON $\rightarrow$ OFF $\rightarrow$ ON $\rightarrow$ OFF $\rightarrow$ ON $\rightarrow$ OFF $\rightarrow$ ON $\rightarrow$ OFF $\rightarrow$ ON. All valid transitions for sensor in dining area.
                    \item Sensor (``In the center of the Living room''): OFF. (First mention, assumes was ON or is initial status). Valid.
                    \item No violations.
                \end{itemize}
            \item \textbf{Spatial Coherence:} M014 is Dining. M020 is Living Room. Plausible: person eating in dining, then M020 in living room turns off (meaning they are not in living room, which is consistent with being in dining). Or they were in living room, it turned off, and they are primarily in dining. No impossible jumps.
            \item \textbf{Logical Reasonableness:} Repeated ON/OFF in M014 (Dining) is highly indicative of presence/activity in the dining area, consistent with eating (e.g., sitting, getting up for seconds, shifting). M020 OFF is consistent with not being in the living room.
            \item \textbf{Check A Passed.}
        \end{itemize}

    \item \textbf{Check B:}
        \begin{itemize}[leftmargin=*, itemsep=1pt]
            \item \textbf{Internal Consistency:} Clean, logical. Repeated M014 activation is fine for eating.
            \item \textbf{Alignment:} Strongly aligns with ``eating'' due to sustained M014 (Dining) activity. M020 OFF is auxiliary, confirming focus is not in living room.
            \item \textbf{Diversity:} Focuses on M014. For eating, this is the primary sensor. The sequence is comprehensive for ``being in dining area and eating''.
            \item \textbf{Score 5:} High consistency, strong alignment, appropriate diversity for an activity focused on the dining area.
        \end{itemize}
\end{itemize}
\paragraph{Sample 2} ...
\paragraph{Sample 3} ...
\paragraph{Sample 4} ...\\
...
\paragraph{Scoring Output:}
\begin{verbatim}
[
  {"Activity": "eating", "SampleID": "Sample1", "Score": 5},
  {"Activity": "eating", "SampleID": "Sample2", "Score": 4},
  {"Activity": "eating", "SampleID": "Sample3", "Score": 4},
  {"Activity": "eating", "SampleID": "Sample4", "Score": 4},
  ...
]
\end{verbatim}
\end{tcolorbox}

\newpage
\section{Additional Visualization Results}
\label{app:visual_results}

\subsection{Cross-Training Performance with Confusion Matrices}
\label{app:confusion_matrices}

\begin{figure}[htbp]
    \centering
    \begin{subfigure}[b]{0.49\textwidth}
        \centering
        \includegraphics[width=\textwidth]{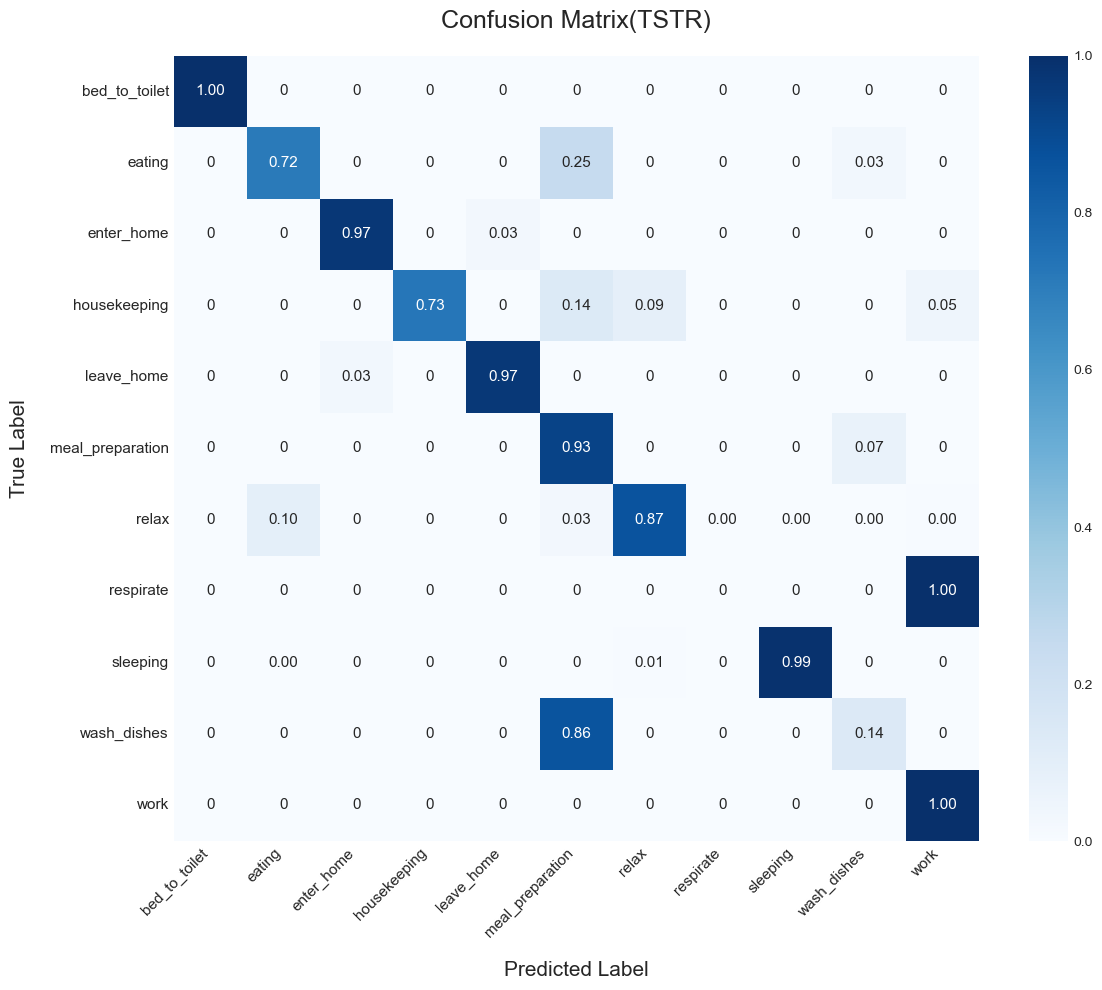}
        \caption{TSTR: Trained on synthetic, tested on real}
        \label{fig:tstr_confusion}
    \end{subfigure}
    \hfill
    \begin{subfigure}[b]{0.49\textwidth}
        \centering
        \includegraphics[width=\textwidth]{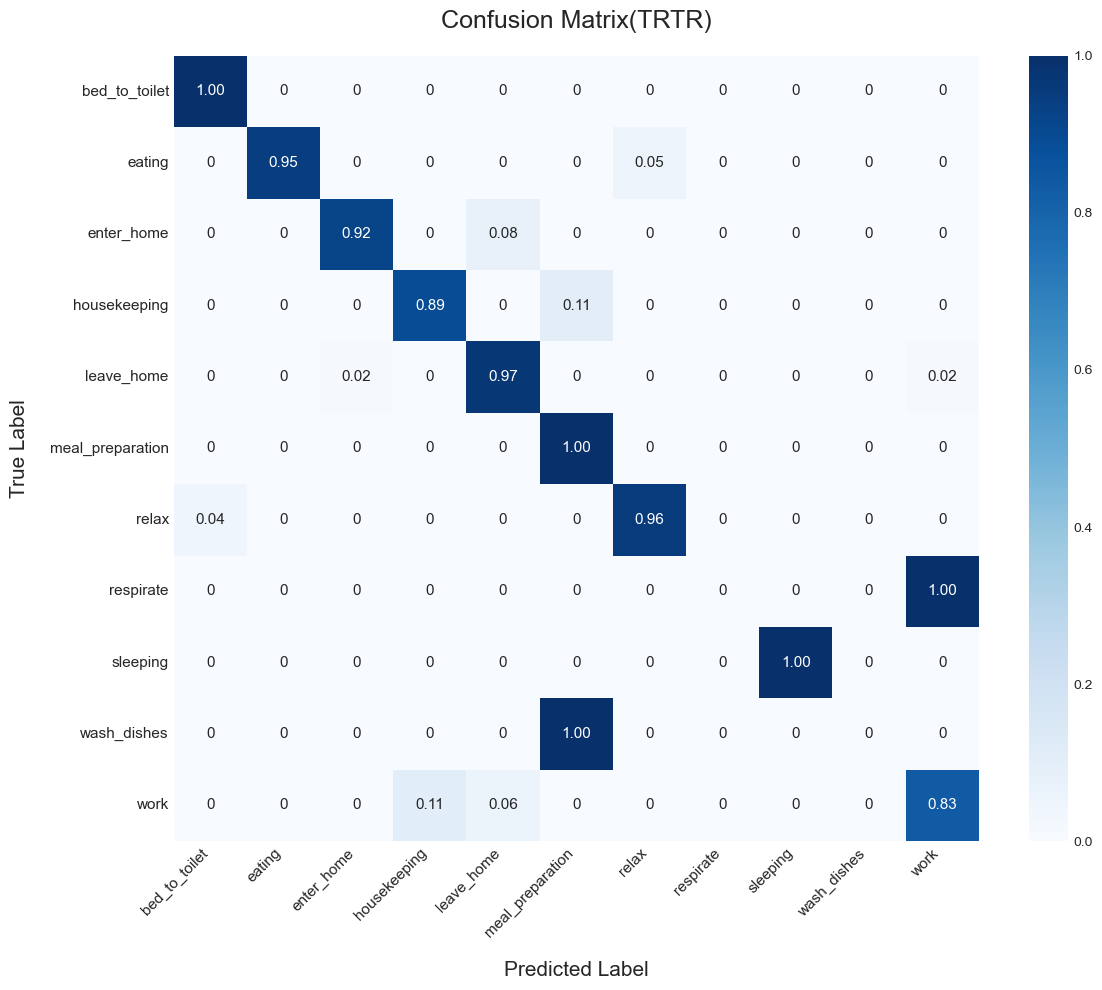}
        \caption{TRTR: Trained on real, tested on real}
        \label{fig:trtr_confusion}
    \end{subfigure}
    \caption{Confusion matrices comparing cross-domain generalization. (a) Shows the classifier trained exclusively on ADLGen-generated synthetic data and tested on real data, demonstrating the synthetic data's utility. (b) Shows the benchmark classifier trained and tested on different splits of real data, representing the ideal performance achievable with available real data.}
    \label{fig:confusion_matrices}
\end{figure}

The comparative analysis of confusion matrices provides critical insights into the generalizability of our synthetic data. Figure \ref{fig:confusion_matrices} presents two confusion matrices side-by-side: Train-on-Synthetic-Test-on-Real (TSTR) and Train-on-Real-Test-on-Real (TRTR). The striking similarity between these matrices demonstrates that classifiers trained exclusively on ADLGen-generated data can achieve discriminative capabilities comparable to those trained on real data, even for minority classes where data scarcity typically hinders recognition performance.

\subsection{Embedding Space Visualization with t-SNE}
\label{app:tsne_visual}
\begin{figure}[ht]
    \centering
     \includegraphics[width=\textwidth]{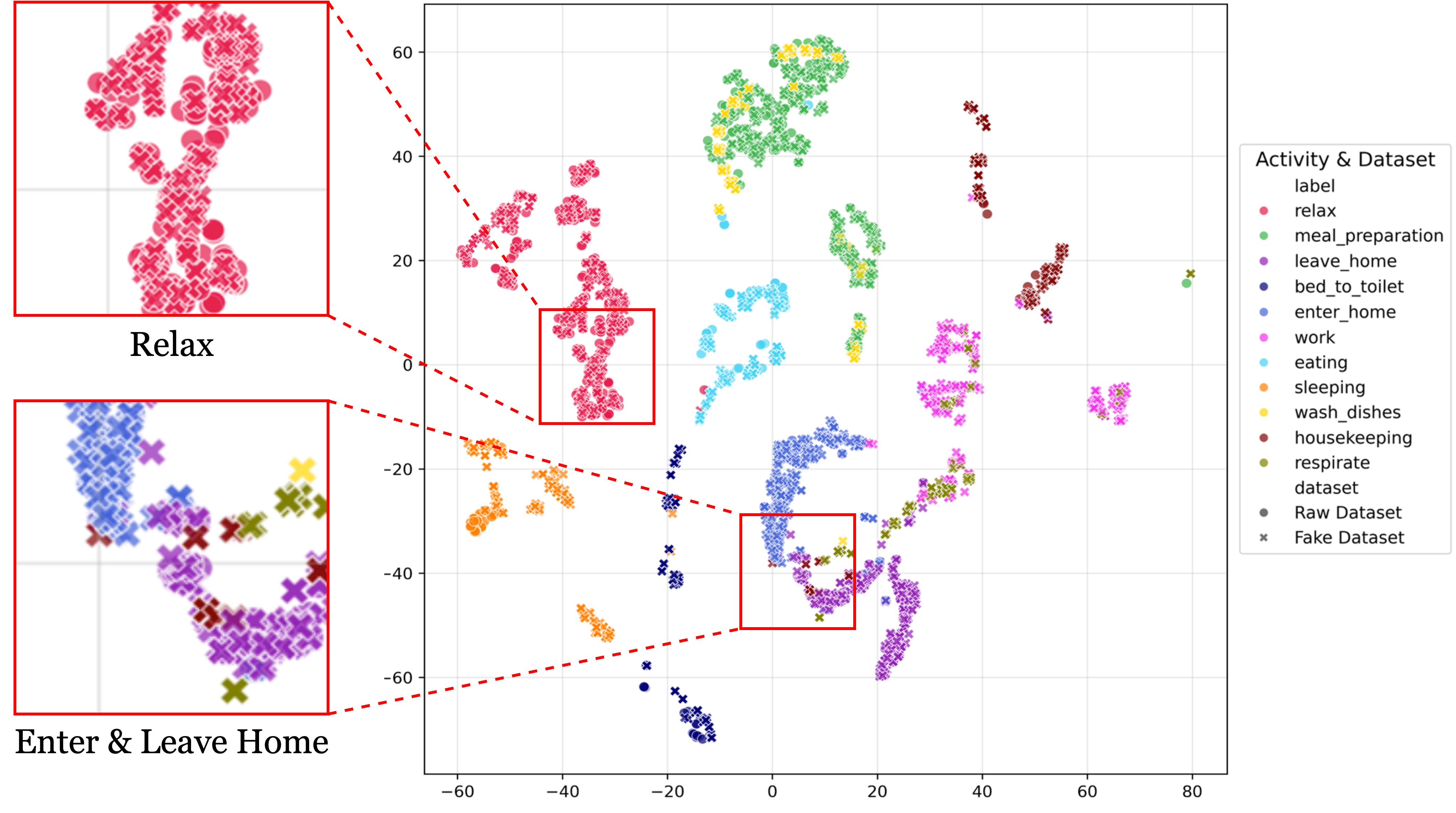}
    \caption{t-SNE visualizations of data embeddings from the Bi-LSTM classifier. Both plots co-visualize. Train on real dataset.}
    \label{tsne_tr}
\end{figure}

\begin{figure}[ht]
    \centering
     \includegraphics[width=\textwidth]{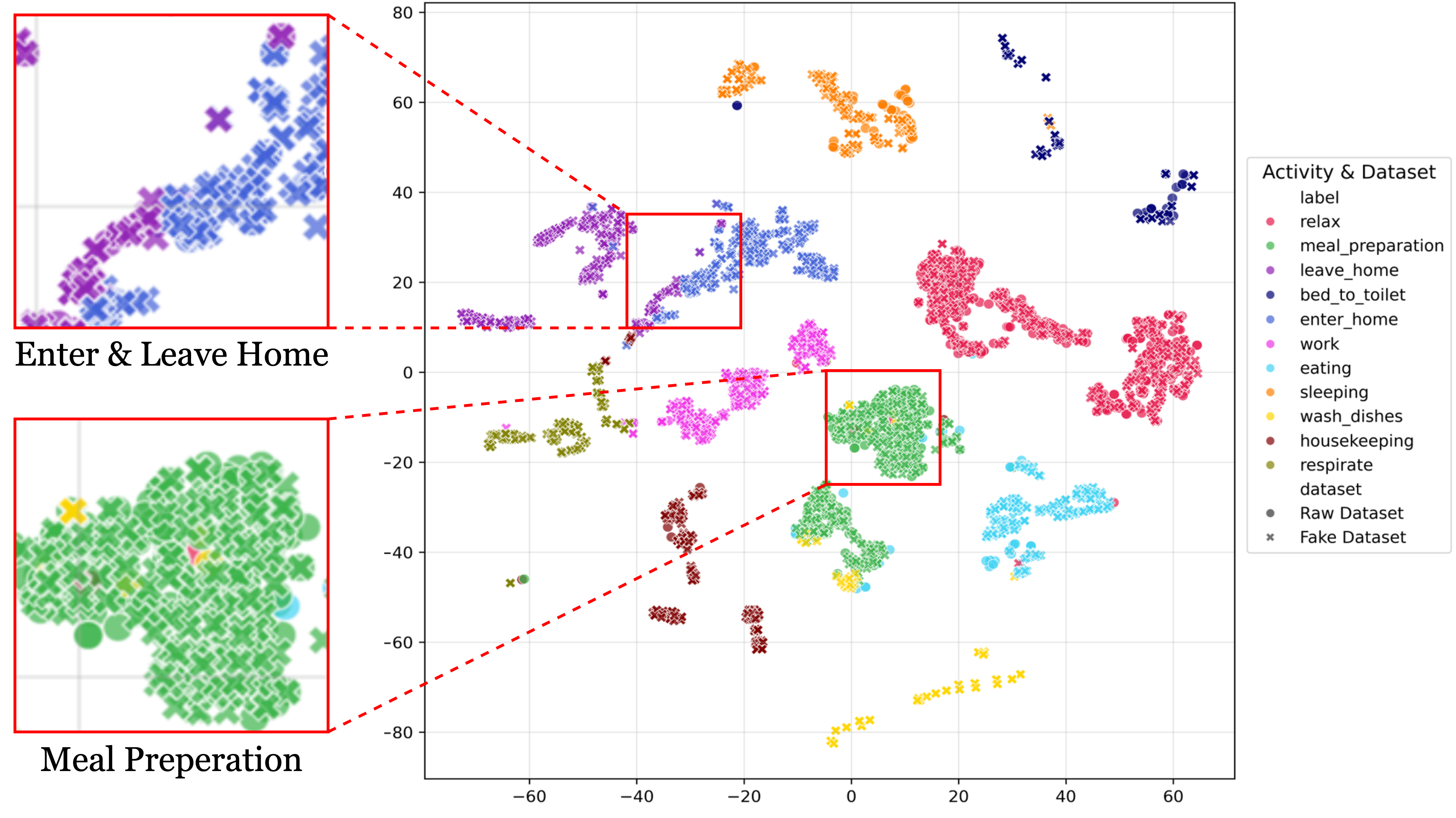}
    \caption{t-SNE visualizations of data embeddings from the Bi-LSTM classifier. Both plots co-visualize. Train on synthetic dataset.}
    \label{tsne_ts}
\end{figure}

To understand how well ADLGen-generated data aligns with real data from a discriminative model's perspective, we visualize their representations using t-SNE. Specifically, we aim to see if classifiers trained on one type of data (e.g., real) can map the other type (e.g., ADLGen-generated) into coherent and corresponding regions in their learned feature space. We employ the same activity recognition classifier architecture (Bi-LSTM~\cite{zhou2016bilstm}) used for downstream utility tests. Embeddings are extracted from an intermediate layer of this classifier prior to the final classification layer.

We generate t-SNE visualizations under two primary classifier training conditions, with results presented in Figure~\ref{tsne_tr} and Figure~\ref{tsne_ts}:
\begin{enumerate}[topsep=0pt,itemsep=0.3em,parsep=0pt,leftmargin=*]
    \item \textbf{Classifier Trained on Real Data (Figure~\ref{tsne_tr}):}
    First, the classifier is trained exclusively on the real Aruba training dataset. We then feed both the real test set and a representative set of ADLGen-generated sequences through this real-trained classifier to obtain their respective embeddings. This allows us to observe how a model optimized for real data perceives the ADLGen-generated distribution relative to real instances.
    \item \textbf{Classifier Trained on ADLGen-Generated Data (Figure~\ref{tsne_ts}):}
    Second, a classifier instance is trained solely on ADLGen-generated data. Subsequently, we pass the real test set and a new set of ADLGen-generated sequences through this ADLGen-trained classifier. This scenario reveals whether a model trained on purely synthetic data learns representations that can effectively structure and separate unseen real-world activities.
\end{enumerate}
For both conditions, the embeddings from real and ADLGen-generated data (processed by the respective single trained classifier) are visualized \textbf{together} on the same t-SNE plot. On these plots, activity classes are consistently colored (e.g., `Relax' is red), while the source dataset is distinguished by marker shape (e.g., circles for real data instances, 'X' markers for ADLGen-generated instances).

\para{Observations}
Figure~\ref{tsne_tr} presents t-SNE visualizations of embeddings derived from a classifier trained on real data. Real data instances (circles) form distinguishable clusters corresponding to their activity labels, with closely related activities positioned proximally in the embedding space. The critical observation is how ADLGen-generated synthetic instances ('X' markers) integrate with these real-data clusters. As highlighted in the zoomed panels, synthetic samples significantly overlap with real samples for both "Relax" and "Enter \& Leave Home" activities. This coherent integration indicates that our synthetic data captures the essential discriminative features of real activities, suggesting high distributional similarity. However, we also observe sub-clustering within activity classes (rather than one homogeneous cluster per activity), reflecting natural variations in how activities are performed—a pattern consistently observed in both real and synthetic data.

Figure~\ref{tsne_ts} examines the feature space learned by a classifier trained exclusively on ADLGen synthetic data. Here, synthetic instances ('X' markers) form clear, separable clusters by activity type, demonstrating the internal consistency of the generated data. Remarkably, when real test data (circles) is projected into this synthetic-trained embedding space, it aligns closely with the corresponding synthetic clusters. As illustrated in the zoomed regions, real "Enter \& Leave Home" instances naturally position themselves within the synthetic cluster of the same activity, and real "Meal Preparation" samples similarly align with their synthetic counterparts. 

We observe two notable phenomena in these visualizations. First, certain synthetic activity classes, particularly "respirate" and "wash dishes" show poorer alignment with their real counterparts. This misalignment likely stems from data limitations (these classes have very few examples in the original dataset) and their similarity to other activities, causing the generative model to learn potentially distorted representations. Second, while synthetic samples overlap significantly with real data clusters, they also expand the distribution boundaries of each class. This expansion suggests that our generative approach enhances the diversity of the feature space, potentially improving generalization by exposing the model to a wider range of valid activity variations. Such potentials are also aligned with the results in experiment~\ref{gernerlization}.

These t-SNE visualizations provide compelling evidence of the bidirectional alignment between our ADLGen-generated sequences and real-world activity data. The consistent co-clustering of same-class instances across different sources (real vs. synthetic) and training regimes (real-trained vs. synthetic-trained) demonstrates that our approach successfully captures the underlying distributional characteristics of smart home activities, validating its potential for data augmentation and transfer learning applications. The observed distribution expansion further reinforces the value of synthetic data for enhancing model robustness in real-world deployment scenarios.

\section{Additional Experimental Results}
\label{app:extra_results}

\subsection{Per-Class Metrics}
\begin{table}[ht]
\centering
\caption{Detailed Per-Class Performance Metrics Across Training-Testing Configurations}
\label{tab:detailed_metrics}
\resizebox{\textwidth}{!}{%
\begin{tabular}{l|cccc|cccc|cccc|cccc}
\toprule
\multirow{2}{*}{\textbf{Class}} & \multicolumn{4}{c|}{\textbf{Synthetic $\rightarrow$ Synthetic}} & \multicolumn{4}{c|}{\textbf{Real $\rightarrow$ Synthetic}} & \multicolumn{4}{c|}{\textbf{Synthetic $\rightarrow$ Real}} & \multicolumn{4}{c}{\textbf{Real $\rightarrow$ Real}} \\
\cmidrule(lr){2-5} \cmidrule(lr){6-9} \cmidrule(lr){10-13} \cmidrule(lr){14-17}
& Acc & Prec & Rec & F1 & Acc & Prec & Rec & F1 & Acc & Prec & Rec & F1 & Acc & Prec & Rec & F1 \\
\midrule
bed\_to\_toilet & 1.00 & 1.00 & 1.00 & 1.00 & 1.00 & 0.99 & 1.00 & 1.00 & 1.00 & 1.00 & 1.00 & 1.00 & 0.95 & 1.00 & 0.95 & 0.98 \\
eating & 1.00 & 0.98 & 1.00 & 0.99 & 0.94 & 1.00 & 0.94 & 0.97 & 0.75 & 0.39 & 0.70 & 0.50 & 0.97 & 1.00 & 0.97 & 0.99 \\
enter\_home & 0.99 & 1.00 & 0.99 & 0.99 & 0.98 & 0.96 & 0.98 & 0.97 & 0.97 & 0.97 & 0.97 & 0.97 & 0.93 & 0.98 & 0.93 & 0.96 \\
housekeeping & 0.97 & 0.98 & 0.97 & 0.98 & 0.93 & 0.84 & 0.93 & 0.88 & 0.96 & 0.94 & 0.62 & 0.74 & 1.00 & 1.00 & 1.00 & 1.00 \\
leave\_home & 0.98 & 0.95 & 0.98 & 0.97 & 0.98 & 0.72 & 0.98 & 0.83 & 0.96 & 0.97 & 0.96 & 0.97 & 0.97 & 0.93 & 0.97 & 0.95 \\
meal\_preparation & 1.00 & 1.00 & 1.00 & 1.00 & 1.00 & 0.51 & 1.00 & 0.67 & 0.97 & 0.88 & 0.92 & 0.90 & 1.00 & 0.97 & 1.00 & 0.98 \\
relax & 0.98 & 1.00 & 0.98 & 0.99 & 1.00 & 0.97 & 1.00 & 0.99 & 0.98 & 0.99 & 0.87 & 0.93 & 1.00 & 1.00 & 1.00 & 1.00 \\
respirate & 0.94 & 1.00 & 0.94 & 0.97 & 0.00 & 0.00 & 0.00 & 0.00 & 0.00 & 0.00 & 0.00 & 0.00 & 0.00 & 0.00 & 0.00 & 0.00 \\
sleeping & 1.00 & 1.00 & 1.00 & 1.00 & 0.99 & 1.00 & 0.99 & 1.00 & 0.99 & 0.99 & 0.99 & 0.99 & 1.00 & 0.98 & 1.00 & 0.99 \\
wash\_dishes & 1.00 & 1.00 & 1.00 & 1.00 & 0.00 & 0.00 & 0.00 & 0.00 & 0.14 & 0.07 & 0.14 & 0.09 & 0.00 & 0.00 & 0.00 & 0.00 \\
work & 1.00 & 0.95 & 1.00 & 0.97 & 0.98 & 0.61 & 0.98 & 0.75 & 0.98 & 0.88 & 0.98 & 0.93 & 0.95 & 0.90 & 0.95 & 0.93 \\
\bottomrule
\end{tabular}%
}
\end{table}

The comprehensive evaluation across different training and testing configurations provides critical insights into our synthetic data quality and utility. We analyze four key configurations: Synthetic→Synthetic (TSTS), Real→Synthetic (TRTS), Synthetic→Real (TSTR), and Real→Real (TRTR), with \textbf{TSTR} and \textbf{TRTS} being particularly informative about synthetic data quality:

\begin{enumerate}[topsep=0pt,itemsep=0.3em,parsep=0pt,leftmargin=*]
    \item \textbf{Synthetic-to-Real (TSTR) Performance:} The TSTR configuration (models trained on synthetic data and tested on real data) achieves an overall F1 score of 0.729 compared to TRTR's 0.797, reaching approximately 91.5\% of real-data performance. This strong alignment indicates our synthetic data effectively captures key discriminative features of real activities. In synthetic data evaluation, TSTR performance closely approaching (e.g., 90\%) TRTR is considered indicative of high-quality synthetic data that successfully bridges the domain gap between synthetic and real distributions.
    
    \item \textbf{Real-to-Synthetic (TRTS) Generalization:} The TRTS configuration shows how well real-data-trained models recognize synthetic activities, essentially measuring how convincingly "real" our synthetic data appears. The strong performance here (with many classes exceeding 0.90 F1) confirms that ADLGen-generated activities closely mimic real data patterns and structures.
    
    \item \textbf{Rare Activity Performance:} Both real and synthetic training data struggle with extremely rare activities like "respirate" (only 6 instances in the dataset), indicating an inherent challenge rather than a limitation specific to synthetic data.
    
    \item \textbf{Class-wise Stability:} The TSTR configuration shows consistent performance across most activity classes, maintaining an F1 score above 0.90 for 7 out of 11 classes, closely matching the TRTR configuration's class distribution.
    
    \item \textbf{Performance on Minority Classes:} For several minority classes (e.g., "work" and "housekeeping"), the synthetic training data delivers competitive or even superior performance compared to real training data, highlighting ADLGen's effectiveness in addressing class imbalance.
\end{enumerate}

The close performance between TSTR and TRTR configurations, coupled with strong TRTS results, demonstrates that our synthetic data not only effectively trains models for real-world deployment but also authentically replicates the statistical properties of real smart home activities. These findings strongly support our hypothesis that generative approaches can effectively address the data scarcity challenges in human activity recognition, particularly for rare activity classes that traditionally suffer from insufficient training examples.

\subsection{Generalizability Across Different Floorplans}
\label{gernerlization}
In this section, we evaluate the generalizability of synthetic data from our approach across different smart home layouts. Specifically, we examine the effectiveness of using synthetic data generated by our ADLGen framework to augment the training data for cross-home Human Activity Recognition (HAR) tasks.

\begin{table}[htbp]
  \centering
  \small
  \caption{Textual sensor-event representations used in the cross–floor-plan experiments}
  \label{tab:sensor_representation}
  \begin{tabular}{
      @{}c                                   
      c                                   
      c                                   
      c                                   
      c                                   
      >{\centering\arraybackslash}p{5.2cm}   
      @{}
  }
    \toprule
    \textbf{Sensor} & \textbf{Location} & \textbf{Time} & \textbf{Type} & \textbf{Value} & \textbf{Textual Representation} \\
    \midrule
    M011 & Home entrance & Morning   & Motion & ON   & Morning: motion sensor at the entrance fired (\textsc{on}) \\
    M015 & Kitchen       & Afternoon & Motion & ON   & Afternoon: motion sensor in the kitchen fired (\textsc{on}) \\
    D002 & Back door     & Evening   & Door   & OPEN & Evening: door sensor at the back door fired (\textsc{open}) \\
    \bottomrule
  \end{tabular}
\end{table}

\firstpara{Setup.}
We use the TDOST-Temporal~\cite{thukral2025tdost} textual representation method (as shown in Table~\ref{tab:sensor_representation}) to ensure fair comparison with the baseline approaches. This representation converts raw sensor events into natural language descriptions while preserving semantic information, which allows our cross-modal approach to align textual descriptions with corresponding spatial representations while maintaining consistency with previous work. 

We conducted a comparative study focusing on the Aruba $\leftrightarrow$ Milan~\cite{cook2012casas} (two biggest dataset in CASAS database) transfer scenarios, which involves two smart homes with different layouts, sensor arrangements, and activity patterns. Figure~\label{fig:laymilan} and Figure~\label{fig:layaruba} show the floorplan sensor layouts of Aruba and Milan dataset. Our evaluation compares two approaches:
\begin{enumerate}[topsep=0pt,itemsep=0.3em,parsep=0pt,leftmargin=*]
    \item \textbf{Baseline (TDOST on Raw dataset)}: The layout-agnostic HAR approach proposed by Thukral et al. was applied to the original raw dataset.
    \item \textbf{Our Approach (TDOST on Raw dataset+ADLGen)}: The same TDOST-Temporal method, but applied on our augmented dataset, where we use ADLGen to enhance the original dataset to create a more balanced training set.
\end{enumerate}

\begin{figure}[h]
  \centering
  \begin{minipage}{0.45\textwidth}
    \centering
    \includegraphics[width=0.8\linewidth]{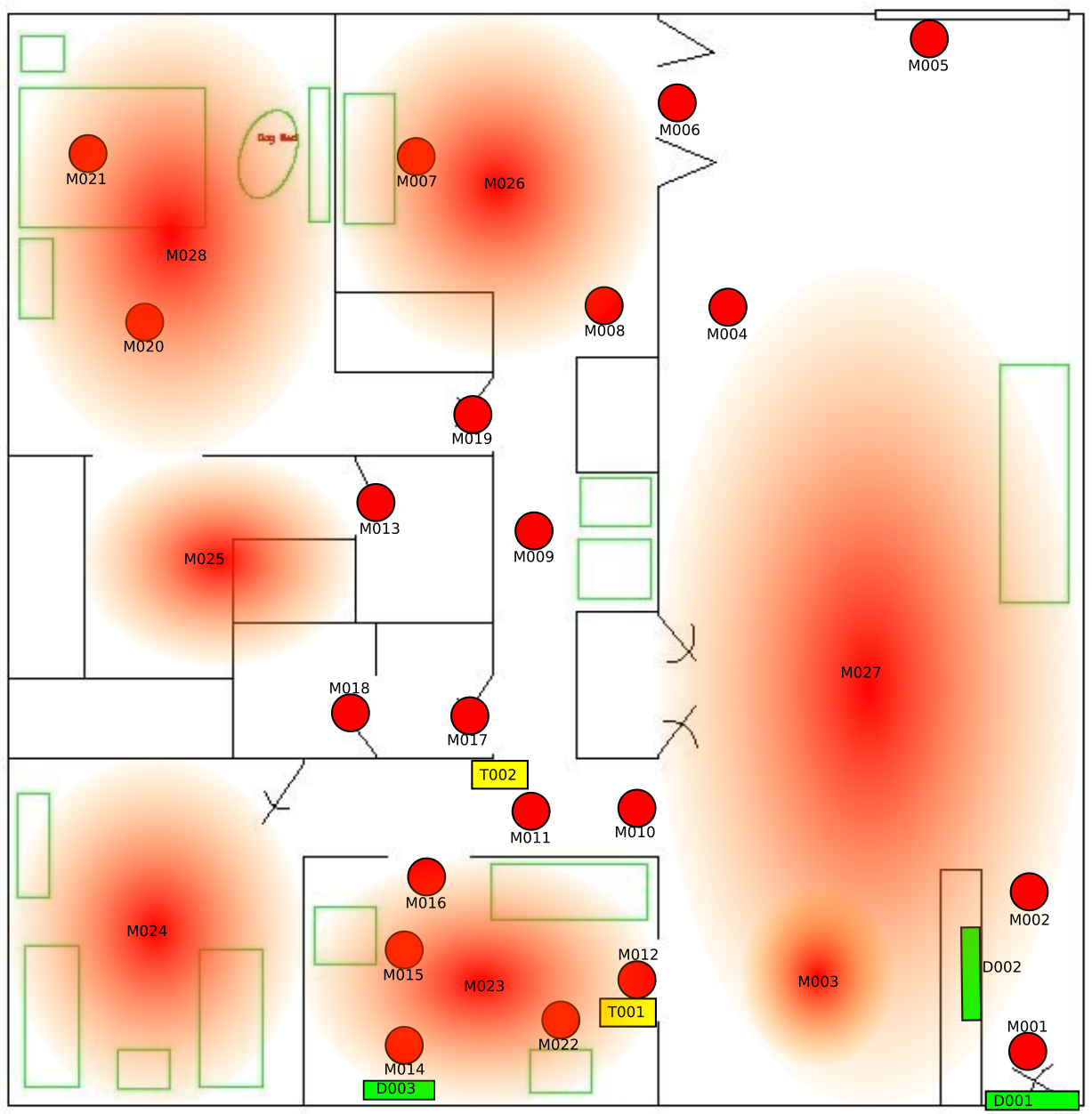}
    \caption{Milan Floorplan}
    \label{fig:laymilan}
  \end{minipage}
  \hfill
  \begin{minipage}{0.45\textwidth}
    \centering
    \includegraphics[width=\linewidth]{fig/aruba.png}
    \caption{Aruba Floorplan}
    \label{fig:layaruba}
  \end{minipage}
\end{figure}

To ensure fair comparative analysis between both approaches, we established a rigorous experimental protocol with standardized conditions across all evaluations. Our methodology employed identical TDOST-Temporal templates for the conversion of raw sensor data into structured textual descriptions, maintaining consistency in the feature extraction phase. The classification architecture remained constant throughout all experiments, utilizing the same sentence transformer coupled with a Bi-LSTM classifier framework to eliminate architectural variations as potential confounding factors. For cross-dataset validity, we carefully mapped activities from both Aruba and Milan datasets to a common standardized activity taxonomy, restricting our evaluation exclusively to the intersection of activity classes present in both environments. Statistical rigor was preserved through implementation of standard cross-validation procedures with consistent train/test partitioning across all experimental conditions, ensuring the reliability and reproducibility of our comparative results.

\begin{table}[h]
\centering
\caption{Cross-layout ADL classification comparisons}
\label{tab:cross_home_results}
\begin{tabular}{lcc}
\hline
\textbf{Method} & \textbf{Accuracy (\%)} & \textbf{F1 Score (\%)} \\
\hline
Raw Aruba $\Rightarrow$Milan (baseline) & 64.2 & 59.8 \\
Aruba+ADLGen$\Rightarrow$Milan (ours) & \textbf{70.5} & \textbf{66.7} \\
\hline
Raw Milan (baseline)$\Rightarrow$Aruba & 61.8 & 53.9 \\
Milan+ADLGen$\Rightarrow$Aruba (ours) & \textbf{68.6} & \textbf{61.3} \\
\hline
\end{tabular}
\end{table}

\para{Results and Discussion}
As shown in table~\ref{tab:cross_home_results}, our approach using TDOST with ADLGen-augmented data achieves approximately 6 to 7 percentage points improvement in both accuracy and F1 score compared to using TDOST with the original raw dataset. This demonstrates that augmenting the source dataset with high-quality synthetic data leads to more robust and generalizable HAR models that transfer better across different home layouts, even when using the same method (TDOST-Temporal) for textual representation.

The performance improvement can be attributed to several factors:
\begin{itemize}
    \item \textbf{Increased pattern diversity}: The synthetic data captures a wider range of sensor activation patterns for each activity, enhancing the model's ability to recognize different manifestations of the same activity.
    \item \textbf{Improved spatial-temporal understanding}: By generating diverse yet realistic activity sequences, ADLGen helps the model build more robust representations of activities that are less dependent on specific sensor layouts.
    \item \textbf{Better generalization to unseen environments}: The augmented training data exposes the model to more variations of activities, reducing overfitting to source-specific patterns.
\end{itemize}

These results preliminarily validate our hypothesis that high-quality synthetic data generation can improve cross-home generalizability for ADL recognition systems, mitigating one of the key challenges in deploying ambient intelligence solutions in new environments without extensive retraining or adaptation.

It's important to note that both approaches follow the same principle of only evaluating on activity classes common to both environments after mapping to standardized activities, ensuring fair comparison. The key difference is the quality and quantity of training data available to the model before transfer.

These findings highlight the potential of synthetic data generation as a complementary approach to layout-agnostic HAR methods. While TDOST provides a powerful framework for handling different layouts through textual abstraction, our ADLGen augmentation addresses the data scarcity and imbalance issues that can limit transfer performance. The combination of these approaches offers a promising pathway toward truly generalizable HAR systems that can be deployed in new environments with minimal adaptation requirements.

\section{Limitations of the Work}
\label{sec:limitations_appendix}
Despite the promising results, our work has several limitations that point to valuable directions for future research. First, our current model does not explicitly incorporate the spatial topology of the environment. While our sign-based encoding implicitly captures some spatial relationships through sensor IDs, we do not directly model the structural information of the floorplan, such as room types (bedroom, kitchen, etc.) or spatial adjacency. Future work could integrate explicit spatial encodings or graph-based representations of sensor locations to enhance the physical plausibility of generated sequences.

Second, the current generation paradigm is limited to synthesizing sequences based on activity classes and optional partial sequences as prefix-conditioning. Extending the model to condition on additional contextual information at inference time, such as activity duration or starting location, could significantly improve flexibility and practical applicability.

Additionally, while our LLM-based evaluation framework provides valuable semantic assessment, it represents an indirect proxy for measuring sequence quality. Developing more direct, domain-specific metrics that can be efficiently computed without requiring translation to natural language remains an important challenge. Future work could explore self-supervised approaches that learn to evaluate sequence quality directly from the sensor event representation space.

Finally, our approach could benefit from incorporating multi-modal data sources beyond binary sensor events, such as continuous environmental sensors (temperature, light) or wearable device data. This integration would enable more comprehensive modeling of human activities and their environmental contexts, potentially leading to more realistic and diverse synthetic ADL data generation.

\section{Societal Impacts}
\label{app:societal_impacts}

Our ADLGen framework for synthetic activity data generation presents both significant benefits and potential concerns that merit thoughtful consideration.

\subsection{Positive Societal Impacts}

The primary societal benefit of our generative approach is enhanced privacy protection in ambient assistive environments~\cite{alrubaie2019privacy}. By generating high-quality synthetic data that closely mimic real sensor activations, ADLGen significantly reduces the need for invasive data collection that captures sensitive behavioral patterns in private spaces. Synthetic data produced by ADLGen also facilitate the development of more robust human activity recognition systems~\cite{li2020activitygan}, particularly for rare activities or underrepresented populations where data collection is challenging. Our cross-domain evaluation demonstrates that models trained on synthetic data approach the performance of those trained on real data, suggesting that synthetic data can effectively support reliable healthcare monitoring systems without compromising quality~\cite{ganin2015unsupervised}.

ADLGen addresses significant data scarcity challenges through democratization of access to training data for researchers who may otherwise lack resources for extensive data collection campaigns. Our ablation studies confirm the effectiveness of our approach in maintaining semantic coherence and physical plausibility, ensuring that the synthetic data are genuinely useful for downstream applications. Additionally, by reducing the need for extensive physical sensor deployments, our approach offers a more environmentally sustainable pathway to developing activity recognition systems, particularly relevant as the field moves toward larger-scale deployments across diverse environments.

\subsection{Potential Negative Impacts}

Despite rigorous evaluation protocols, any synthetic data generation system carries the risk of producing sequences that misrepresent reality in subtle ways~\cite{jordon2018measuring}. If models trained on such data are deployed in critical healthcare applications, these misrepresentations could lead to incorrect inferences about human behavior. Our LLM-based refinement pipeline specifically addresses potential inconsistencies, although ongoing validation against real-world data remains necessary. Furthermore, generative models can inadvertently amplify biases present in training data~\cite{buolamwini2018gender}, potentially leading to disparate performance across different demographic groups in healthcare monitoring applications.

Although our work is motivated by healthcare care applications, the technology could potentially be adapted to enhance surveillance capabilities, possibly leading to systems that predict behavior in ways that infringe on privacy or civil liberties. Furthermore, the computational resources required to implement our generative framework may not be universally accessible, potentially exacerbating divides between well-resourced and under-resourced research groups. Finally, there is a risk that the availability of high-quality synthetic data might discourage necessary real-world data collection, though we emphasize that our approach should complement, rather than replace, thoughtful real-world data collection.

\subsection{Recommended Practices}

To maximize benefits while mitigating risks, we recommend several practices~\cite{jobin2019global}: (1) transparency about the usage of synthetic data in activity recognition systems; (2) regular validation against diverse real-world data sources; (3) thorough bias audit of both input and generated data; (4) robust privacy protections when integrating synthetic and real data; and (5) appropriate ethical review for applications in sensitive contexts. By paying careful attention to these considerations, ADLGen can make significant positive contributions to healthcare and assistive technologies while minimizing potential negative impacts.

\newpage

\newpage
\bibliography{ref}
\bibliographystyle{unsrt}

\end{document}